\documentclass[preprint,12pt]{elsarticle}
\usepackage[margin=1in]{geometry}
\usepackage{amsmath,amssymb,amsthm}
\usepackage{graphicx}
\usepackage{booktabs}
\usepackage{multirow}
\usepackage[section]{placeins}
\usepackage{algorithm}
\usepackage{algpseudocode}
\usepackage{xcolor}

\journal{Engineering Structures}

\begin{document}
\begin{frontmatter}

\title{Physics-Guided Dual-Stream Heterogeneous Graph Neural Network for Predicting Full-Field Structural Response of Stiffened Panels}

\author[1]{Yuecheng Cai}
\author[1,2]{Jasmin Jelovica\corref{cor1}}
\cortext[cor1]{Corresponding author}
\ead{jasmin.jelovica@ubc.ca}

\address[1]{Department of Mechanical Engineering, The University of British Columbia, Vancouver, BC V6T 1Z4, Canada}
\address[2]{Department of Civil Engineering, The University of British Columbia, Vancouver, BC V6T 1Z4, Canada}

\begin{abstract}
Iterative design and optimization of large, complex structures require fast and accurate prediction of stress, displacement, and other fields. Finite element analysis (FEA) is computationally expensive for this task. Existing neural network surrogates often struggle with varying topologies and complex boundary conditions. This study proposes the novel Dual-Stream Heterogeneous Graph Neural Network (DS-HGNN) for full-field stress and displacement prediction in thin-walled structures, demonstrated on box beams made of stiffened panels. DS-HGNN operates on panel-level heterogeneous graph representations and introduces physics-guided edge states initialized from edge types, spatial information, and boundary kinematics. These states are updated through dual-stream message passing that separates longitudinal and transverse structural information while allowing cross-stream exchange. Geometry and loading effects are incorporated through Feature-wise Linear Modulation (FiLM)-conditioned 1-D spectral convolutions, and physical fields are reconstructed using a spectral--bypass low-rank readout. The model is evaluated on stiffened panel datasets with different geometries, boundary kinematics, loading conditions, and material nonlinear responses. DS-HGNN achieves the lowest stress and displacement RMSE compared with six benchmark heterogeneous graph neural network models. It also reaches comparable accuracy to the strongest benchmark models using $19\%$--$38\%$ fewer training samples. A targeted evaluation further shows that DS-HGNN captures yield and post-yield stress features.
\end{abstract}


\end{frontmatter}

\section{Introduction}
Many thin-walled structures in civil, aerospace, and ocean engineering are assembled from stiffened panels, including bridge girders, aircraft wings and fuselages, ship hulls, and offshore structures. Their stress and displacement fields govern strength assessment, fatigue performance, and design compliance under diverse loading and boundary conditions \cite{hughes2005ship,clough1990original,thompson2017ansys}. Stiffened panels are composed of multiple plate units with varying geometries, boundary conditions, and loading conditions, so finite element analysis (FEA) remains the standard tool for computing their stress and displacement fields with sufficient accuracy. However, high fidelity FEA is computationally expensive for large scale structures because accurate simulations need to resolve interactions between structural components, local stress variations, and material nonlinearities. This typically requires fine spatial discretization and iterative nonlinear solvers.

The computational cost of high fidelity FEA is particularly high in iterative design and optimization workflows. Structural optimization explores design spaces with varying geometries, stiffener arrangements, loading conditions, and boundary states. State-of-the-art optimization algorithms can require hundreds of thousands to millions of design evaluations to converge \cite{jelovica2024improved,cai2023neural}, and each evaluation often requires a full nonlinear FEA solve. Repeated high fidelity FEA at this scale is therefore computationally prohibitive. This motivates surrogate models that can predict stress and displacement fields accurately across diverse geometries and loading scenarios.

Traditional data driven surrogates such as multilayer perceptrons (MLPs) and convolutional neural networks (CNNs) have been applied to structural response prediction \cite{hornik1989multilayer,papadrakakis1998structural,mai2022robust,shojaeefard2013modelling,kabir2021failure,sun2021prediction}, but they are often inadequate for stiffened panels because these structures involve varying topologies, heterogeneous components, and localized high gradient stress and displacement responses. Comparative studies on stiffened panels showed that graph neural networks (GNNs) outperform MLPs and CNNs, suggesting that message passing provides useful inductive biases for modeling relational dependence among structural entities and accommodating varying structural topologies \cite{cai2025comparison,mokhtari2025comparison}. More broadly, GNNs have demonstrated strong capability in computational mechanics through their ability to propagate information along physically meaningful connections \cite{gilmer2017neural,hamilton2017inductive,xu2018powerful,battaglia2018relational,lino2021simulating}. In particular, heterogeneous graph neural networks (HGNNs), including relation aware and attention-based variants such as RGCN, HGT, and HAN, extend this capability by distinguishing different kinds of structural entities and their connections \cite{wang2019heterogeneous,hu2020heterogeneous,fu2020magnn,velivckovic2017graph,brody2021attentive}.

Building on graph-based surrogate modeling, our previous studies have progressively advanced stress and displacement field prediction for stiffened panels. In \cite{cai2024efficient}, a homogeneous graph representation was proposed for efficient stress prediction under idealized boundary conditions. This was extended in \cite{cai2026heterogeneous} to a heterogeneous graph representation that accommodates non-uniform boundary kinematics and pressure distributions by introducing different node types for structural geometry, physical edges, loading, and boundary conditions. The heterogeneous representation was subsequently integrated into a hybrid global--local framework for ship hull analysis \cite{cai2026hybrid}. In that framework, the graph surrogate was trained at the panel-level rather than on the entire structural system, enabling the panel-level model to be embedded in larger and more complex structures. These developments demonstrate that graph-based surrogates can represent interactions among structural components while substantially reducing computational cost.

Nevertheless, challenges remain when general-purpose HGNN models are used for thin-walled structures with diverse loading and boundary kinematic inputs. Existing HGNN variants are effective at distinguishing node and relation types through relation-aware aggregation or attention mechanisms \cite{wang2019heterogeneous,hu2020heterogeneous,fu2020magnn}, but their update rules are usually designed as generic message-passing operators rather than as structural response models. For stiffened panels, the relevant relations have specific mechanical roles. Plate-edge interfaces transmit boundary information, while loading and boundary kinematics provide conditioning inputs. The longitudinal and transverse plate directions could have different magnitude and profile of stress and displacement \cite{cai2026heterogeneous}. When geometry, loading, boundary kinematics, and relation direction are merged into a single latent representation before propagation, these roles are learned only implicitly. This limitation motivates the incorporation of inductive biases tailored to thin-walled structures into the surrogate modeling process.

Physics-informed neural networks (PINNs) provide a common way to introduce physical knowledge by adding governing-equation residuals and boundary-condition penalties to the training loss \cite{raissi2019physics,karniadakis2021physics,haghighat2021physics}. However, this loss-based formulation is less natural for the present stiffened panel surrogate problem. The extracted stiffened panels do not correspond to a single fixed PDE problem over a fixed computational domain. The response of stiffened panels is governed by their complex geometry, discrete and spatially varying loading, nonlinear constitutive models, and panel boundary kinematics. Enforcing these effects through PDE, boundary, and interface residuals would require problem-specific loss terms and weighting strategies, and PINN training can become ill-conditioned when competing residual terms are imposed \cite{wang2022and}. Related neural network-based domain decomposition methods address coupling between subdomains through explicit interface penalty losses or matching conditions \cite{jagtap2020conservative,moseley2023finite}, but these treatments introduce additional hyperparameters and may be restrictive for complex assembled geometries. These challenges motivate the incorporation of structural mechanics inductive biases into HGNN-based field prediction models.

These challenges suggest that structural field prediction requires models that incorporate mechanics related information during message passing rather than relying only on generic graph aggregation. In addition, the model needs to reconstruct spatially continuous stress and displacement fields from the learned structural representations. Fourier neural operators (FNOs) learn mappings between function spaces by representing operators in the frequency domain and have shown strong capability for parametric PDE learning and physical field prediction \cite{li2021fourier,kovachki2023neural,li2024physics}. More recently, Lifting Product Fourier Neural Operators (LPFNO) addressed boundary to domain mapping by combining one-dimensional Fourier operator branches with an outer-product lifting step to reconstruct domain fields \cite{kashi2024learning}. Although these spectral operator models provide useful mechanisms for field reconstruction, applying them as standalone neural operators to thin-walled panel structures remains challenging because interactions across shared plate interfaces must be represented explicitly. This motivates using spectral operator ideas as local architectural components within a graph-based model, rather than treating operator learning as the primary formulation.

To bridge this gap, we propose the novel Dual-Stream Heterogeneous Graph Neural Network (DS-HGNN), a physics-guided neural network architecture for stress and displacement field prediction in thin-walled structures such as bridge girders and ship hull girders. These structures can often be decomposed into a large number of interacting plate-like components. In this study, we demonstrate the effectiveness of the proposed approach on three box beam cases composed of stiffened panels with varying geometries, loading conditions, boundary kinematics, and material nonlinear responses. The DS-HGNN design incorporates structural mechanics information into heterogeneous graph learning and enables efficient prediction of full stress and displacement fields. The main contributions of this study are as follows:
\begin{itemize}
    \item We propose a new physics-guided edge state initialization method that encodes structural edge types, spatial information, and boundary kinematic values into edge states for subsequent message passing.
    \item We develop a dual-stream heterogeneous graph message passing mechanism that separately processes field information along the two plate directions while allowing information exchange between the two directions.
    \item We incorporate geometry- and loading-guided conditioning so that structural geometry and loading information can modulate the field prediction for each panel.
    \item We develop a new spectral--bypass readout that combines learnable directional spectral bases with a bypass branch and outer product field reconstruction to recover full stress and displacement fields.
    \item We validate DS-HGNN using stiffened panel samples extracted from box beam analyses through benchmark model comparison, ablation and variant studies, a hyperparameter study, representative panel demonstrations, and targeted evaluation of panels with plastic material response.
\end{itemize}

\section{Preliminaries}
\subsection{Heterogeneous Graph Networks}
We denote a heterogeneous graph by
\begin{equation}
\mathcal{G}=(\mathcal{V},\mathcal{E},\mathcal{T},\mathcal{R}),
\end{equation}
where $\mathcal{V}$ is the node set, $\mathcal{E}$ is the edge set, $\mathcal{T}$ is the node-type set, and $\mathcal{R}$ is the relation-type set. Each node $v\in\mathcal{V}$ has a type mapping $\tau(v)\in\mathcal{T}$, and each directed edge $(u,v)\in\mathcal{E}$, with $u,v\in\mathcal{V}$, has a relation type $r=\rho(u,v)\in\mathcal{R}$. Under this formulation, a heterogeneous graph encodes both connectivity and distinct semantic or physical roles through node types and relation types \cite{wang2019heterogeneous,hu2020heterogeneous,fu2020magnn}.

For each relation type $r\in\mathcal{R}$, we define a relation-specific adjacency matrix
\begin{equation}
A^{(r)}\in\{0,1\}^{|\mathcal{V}|\times|\mathcal{V}|},
\qquad
\left[A^{(r)}\right]_{uv}=\begin{cases}
1, & (u,v)\in\mathcal{E}\ \text{and}\ \rho(u,v)=r,\\
0, & \text{otherwise},
\end{cases}
\end{equation}
in the matrix entry $[A^{(r)}]_{uv}$, the row index $u$ denotes the source node and the column index $v$ denotes the target node under directed message passing. The relation-aware neighborhood of node $v$ is therefore
\begin{equation}
\mathcal{N}_r(v)=\{u\in\mathcal{V}\mid [A^{(r)}]_{uv}=1\}.
\end{equation}

Given this representation, a heterogeneous graph network (HGNN) parameterizes relation-aware message passing over $\mathcal{G}$. A generic update at layer $\ell$ is
\begin{equation}
h_v^{(\ell+1)}=
\phi_{\tau(v)}^{(\ell)}\!\left(
h_v^{(\ell)},
\bigoplus_{r\in\mathcal{R}}
\mathrm{AGG}_r\!\left(
\left\{\psi_r^{(\ell)}(h_u^{(\ell)},h_v^{(\ell)})\,\middle|\,u\in\mathcal{N}_r(v)\right\}
\right)
\right),
\end{equation}
where $h_v^{(\ell)}\in\mathbb{R}^{d_\ell}$ is the hidden state of node $v$ at layer $\ell$ with feature dimension $d_\ell$, $\mathcal{N}_r(v)$ is the set of source neighbors that send messages to $v$ under relation $r$, $\psi_r^{(\ell)}(\cdot)$ is the relation-specific message map, $\mathrm{AGG}_r(\cdot)$ is a permutation-invariant aggregator (e.g., sum, mean, or attention-based aggregation), $\bigoplus$ denotes relation-level fusion (e.g., concatenation or summation), and $\phi_{\tau(v)}^{(\ell)}(\cdot)$ is the node type specific update function. This general formulation covers representative HGNN variants, including relation-aware and attention-based models \cite{velivckovic2017graph,brody2021attentive,wang2019heterogeneous,hu2020heterogeneous}.

For structural mechanics, heterogeneous graphs and HGNNs provide an effective representation of multiple physical entities and interfaces between components in a single graph. They also allow relation aware information propagation, which can help represent interactions among components under non-uniform boundary conditions and loading conditions \cite{pfaff2020learning,lino2022multi,gao2022finite,cai2024efficient}. These properties are particularly important for stiffened panels, where the distinct physical roles of plates, stiffener webs, and flanges, together with their interactions, strongly influence stress and displacement fields.

\subsection{Heterogeneous Graph Representation for Stiffened Panels}
\begin{figure}[!htbp]
\centering
\includegraphics[width=0.9\linewidth]{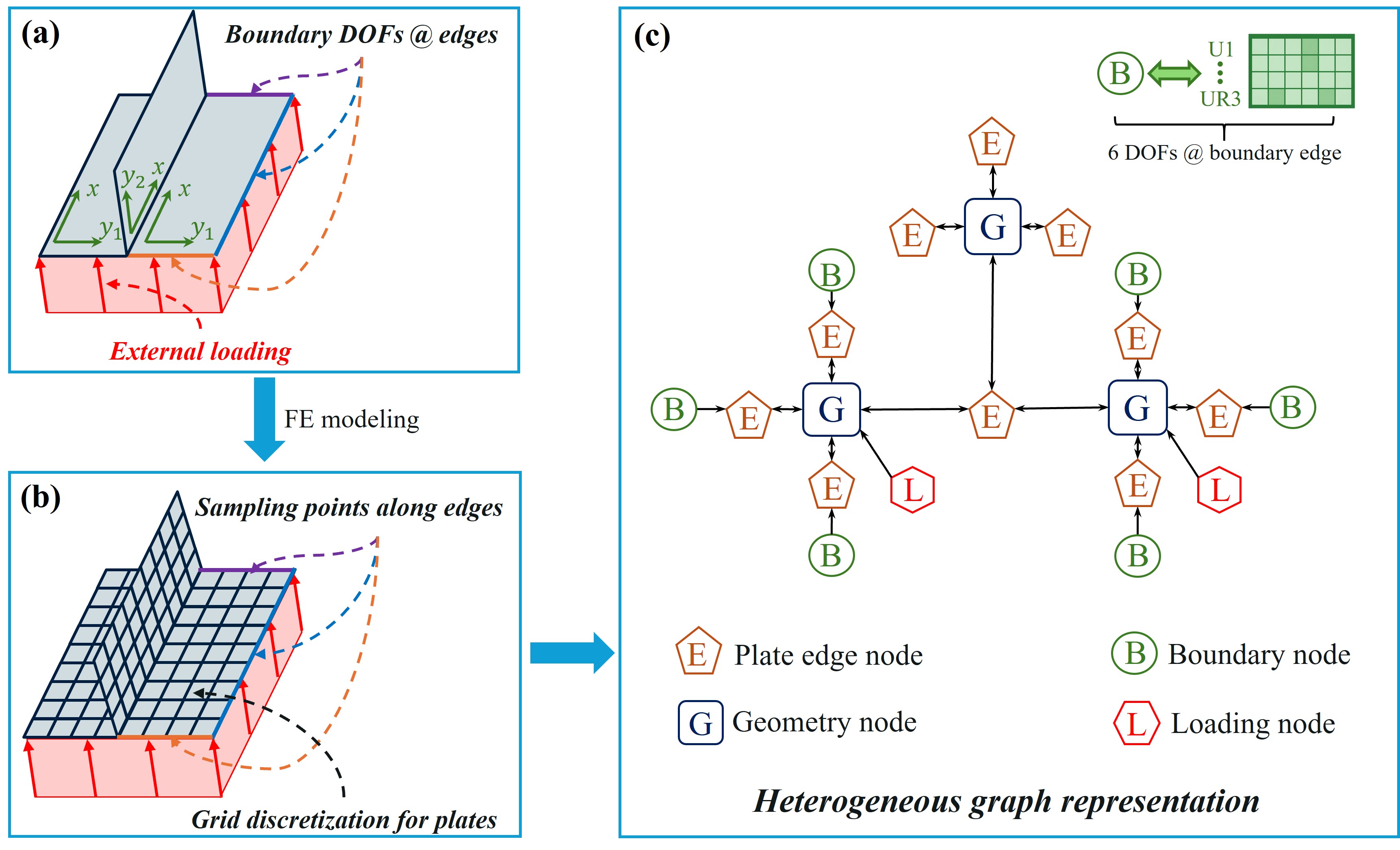}
\caption{Heterogeneous graph representation of a stiffened panel.}
\label{fig:hetero_graph_representation}
\end{figure}

Motivated by the above advantages, our previous work developed a heterogeneous graph representation for stiffened panels subjected to non-uniform boundary kinematics and external loadings \cite{cai2026heterogeneous}. Each stiffened panel is decomposed into multiple plates and discretized as a heterogeneous graph with node partition:
\begin{equation}
\mathcal{V}=\mathcal{V}_{geo}\cup\mathcal{V}_{edge}\cup\mathcal{V}_{loading}\cup\mathcal{V}_{bc}.
\end{equation}
Here, \textit{geo} nodes represent decomposed plate-like subdomains, \textit{edge} nodes represent physical edges of each plate, \textit{loading} nodes encode external loads, and \textit{boundary} nodes encode boundary kinematics applied at each edge. Fig.~\ref{fig:hetero_graph_representation} illustrates an example of this heterogeneous graph representation. For a typical stiffened panel, the raw features are $x_i^{geo}\in\mathbb{R}^{d_{geo}}$, $x_e^{edge}\in\mathbb{R}^{d_{edge}}$, $x_p^{loading}\in\mathbb{R}^{d_{load}}$, and $x_b^{boundary}\in\mathbb{R}^{S\times 6}$, where $S$ is the number of sampling points along a plate boundary and $6$ corresponds to the six boundary degrees of freedom. In this context, boundary kinematics refers to the displacement and rotational degrees of freedom specified at the sampled points along an extracted panel edge.

\FloatBarrier

Different node types are connected by different relation types. Relations between \textit{geo} nodes and \textit{edge} nodes are bidirectional and depend on physical adjacency and boundary orientation. In this representation, each stiffened panel is decomposed into plate components, such as the plate strip between adjacent stiffeners, the stiffener web, and the flange strips connected to the web. For each plate component, the longer in-plane direction is defined as the local longitudinal ($x$) direction, and the in-plane direction perpendicular to it is defined as the local transverse direction. Fig.~\ref{fig:hetero_graph_representation}(a) illustrates the local coordinate systems used for different plate components and their associated edge nodes. The local $x$ direction defines the longitudinal direction, while the local transverse directions, such as $y_1$ for the plate strip and $y_2$ for the stiffener web, may have different orientations. These local transverse directions are grouped into the same transverse relation set. Based on this local plate coordinate definition, the \textit{geo}--\textit{edge} relation types are grouped into two directional sets. Relations in the longitudinal set $\mathcal{R}_x$ connect \textit{geo} nodes to \textit{edge} nodes associated with boundaries along the longer in-plane direction of each plate component. Relations in the transverse set $\mathcal{R}_y$ connect \textit{geo} nodes to \textit{edge} nodes associated with boundaries along the perpendicular in-plane direction. This transverse set covers transverse plate edges as well as stiffener web and flange edges and interfaces. In contrast, \textit{loading} nodes connect to \textit{geo} nodes and \textit{boundary} nodes connect to \textit{edge} nodes through directed, one-way relations, providing conditioning information that injects loading and boundary kinematic data into the graph.

For each \textit{geo} node $i\in\mathcal{V}_{geo}$, the model predicts a field on a uniform canonical grid:
\begin{equation}
\hat{Y}_i\in\mathbb{R}^{n_x\times n_y\times C},
\end{equation}
where $n_x$ and $n_y$ are grid resolutions in the longitudinal and transverse directions, respectively, and $C$ is the number of output channels. In this study, $C=1$ for von Mises stress prediction. For displacement prediction, $C=3$, corresponding to the three translational displacement components. These two targets are selected as the field quantities of interest in this study. This canonical-grid representation decouples the prediction domain from the FE mesh and provides a structured grid for the spectral readout.

\subsection{Spectral Convolutions for Field Reconstruction}
Fourier Neural Operator (FNO) learns mappings between function spaces by parameterizing operators in the spectral domain \cite{li2021fourier,kovachki2023neural}. Given a discretized input function with $N$ spatial points, an FNO first applies a pointwise lifting operator $\boldsymbol{Q}:\mathbb{R}^{m}\to\mathbb{R}^{n_e}$ to project each point from the $m$-dimensional input space into a hidden embedding of dimension $n_e$. The resulting representation $\boldsymbol{h}\in\mathbb{R}^{N\times n_e}$ is then processed through a sequence of FNO layers $\boldsymbol{F}_{\theta_i}:\mathbb{R}^{N\times n_e}\to\mathbb{R}^{N\times n_e}$, each defined as:
\begin{equation}
\boldsymbol{F}_{\theta_i}(\boldsymbol{h})=\sigma\!\left(\mathcal{F}^{-1}\!\left(\boldsymbol{R}\cdot\mathcal{F}(\boldsymbol{h})\right)+\boldsymbol{W}_i\,\boldsymbol{h}\right),
\end{equation}
where $\mathcal{F}$ and $\mathcal{F}^{-1}$ denote the forward and inverse Fourier transforms, $\boldsymbol{R}$ is a learnable linear operator that acts independently on each Fourier mode and couples the $n_e$ embedding channels, $\boldsymbol{W}_i:\mathbb{R}^{n_e}\to\mathbb{R}^{n_e}$ is a pointwise linear map, and $\sigma(\cdot)$ is a nonlinear activation function. This formulation can be effective for long-range spatial coupling and multi-scale field patterns \cite{li2021fourier,kovachki2023neural}.

The Lifting Product Fourier Neural Operator (LPFNO) adapts FNO to boundary-to-domain learning, where the goal is to map a boundary function $\boldsymbol{g}:\Gamma\to\mathbb{R}^{m}$ defined on the lower-dimensional boundary $\Gamma\subset\partial\Omega$ to a solution over the higher-dimensional domain $\Omega$ \cite{kashi2024learning}. LPFNO processes the boundary function through two separate FNO blocks, each consisting of $l$ stacked FNO layers, producing two 1-D hidden representations $\boldsymbol{v}_1,\boldsymbol{v}_2\in\mathbb{R}^{N\times n_e}$:
\begin{equation}
\boldsymbol{v}_k = \left(\prod_{i=1}^{l}\boldsymbol{F}_{\theta_{k,i}}\right)\!\circ\,\boldsymbol{Q}_k\!\left(\boldsymbol{g}\right),\qquad k\in\{1,2\},
\end{equation}
where $\prod$ denotes sequential composition and $\boldsymbol{Q}_k$ is the lifting operator for each branch. The two representations are then combined through a \textit{lifting product} that lifts the 1-D boundary information to a 2-D domain field. The lifting product is applied independently to each embedding channel as an outer product,
\begin{equation}
\boldsymbol{w}[\,:\,,\,:\,,\,j\,] = \boldsymbol{v}_1[\,:\,,\,j\,]\;\boldsymbol{v}_2[\,:\,,\,j\,]^{\!\top},\qquad j=1,\dots,n_e.
\end{equation}
The resulting tensor is $\boldsymbol{w}\in\mathbb{R}^{N\times N\times n_e}$. Because the two directional representations are combined by an outer product, the reconstruction avoids a direct dense mapping from boundary embeddings to all domain grid points. A pointwise projection $\boldsymbol{P}:\mathbb{R}^{n_e}\to\mathbb{R}^{m}$ then maps the lifted embedding back to the physical variable space. By factoring the domain solution into two boundary-derived components, LPFNO provides a structured mechanism for boundary-to-domain field reconstruction while keeping the spectral processing on one-dimensional boundary representations.

In the present work, FNO and LPFNO are used as architectural references rather than as the formulation of the proposed model. Specifically, one-dimensional spectral convolutions motivate the directional latent-sequence updates in DS-HGNN, and the LPFNO lifting-product structure motivates the low-rank outer-product readout in Section~\ref{sec:spectral_bypass_readout}. These components are embedded into a heterogeneous graph neural network whose main computation remains message passing over shared physical interfaces. In this way, spectral processing supports directional basis reconstruction on each plate, while graph message passing represents interactions among adjacent plate components in irregular plate-based structural topologies.

\section{Proposed DS-HGNN Architecture}
In this study, the objective is to predict a physical field tensor $\hat{Y}_i\in\mathbb{R}^{n_x\times n_y\times C}$ for each decomposed plate component of a stiffened panel, represented by a \textit{geo} node $i\in\mathcal{V}_{geo}$ in the heterogeneous graph $\mathcal{G}$. Throughout this section, the term \textit{field} refers to a target physical quantity defined over the corresponding plate component. In this study, it corresponds to either the von Mises stress field ($C=1$) or the displacement field ($C=3$, three translational displacement components), although the architecture could be extended to other scalar or vector field quantities.

\begin{figure}[!htbp]
\centering
\includegraphics[width=\linewidth]{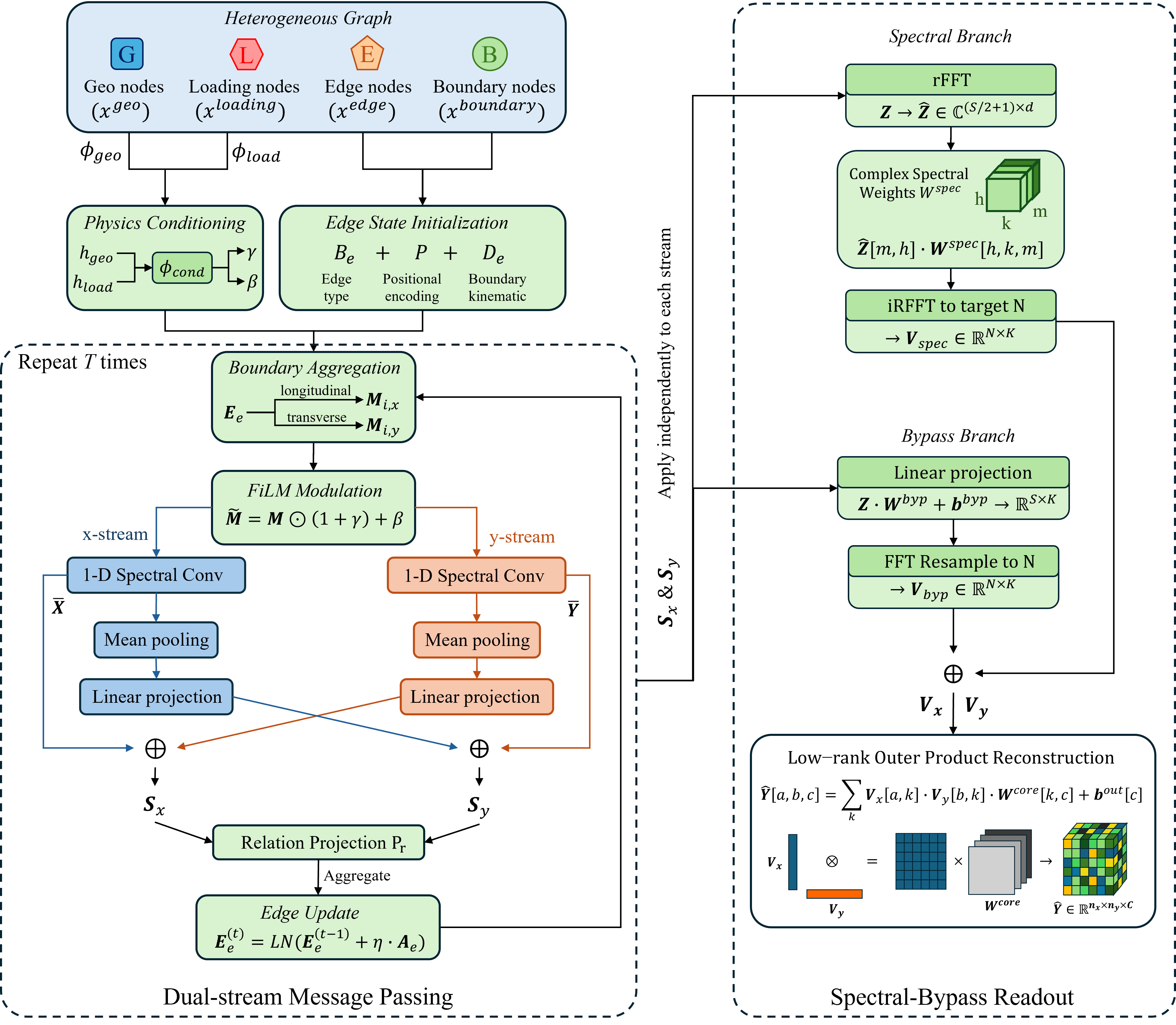}
\caption{Overview of the DS-HGNN architecture. Edge states are first initialized from edge type information, positional encodings, and boundary kinematic values. The initialized edge states are then refined over $T$ dual-stream message passing layers through directional boundary aggregation, FiLM-conditioned modulation, dual-stream spectral convolution with cross-stream crosstalk, and residual edge updates. The refined latent sequences for each plate, $S_{x,i}^{(T)}$ and $S_{y,i}^{(T)}$, are decoded through spectral and bypass branches and reconstructed into the output field $\hat{Y}_i$ via low-rank outer product decoding.}
\label{fig:architecture}
\end{figure}

An overview of the complete architecture is illustrated in Fig.~\ref{fig:architecture}. DS-HGNN consists of four main architectural components: physics-guided edge state initialization, dual-stream message passing with cross-stream crosstalk, geometry- and loading-guided FiLM conditioning, and a spectral--bypass low-rank readout. These components are organized into three operational stages. First, in the \textit{edge state initialization} stage (Section~3.1), each \textit{edge} node is initialized as a latent sequence over predefined boundary sampling points using edge type information, positional information, and boundary kinematic values. Second, in the \textit{iterative message passing} stage (Sections~3.2--3.4), the model applies $T$ message passing layers. Each layer performs relation aware read and write updates between \textit{geo} and \textit{edge} nodes, aggregates edge states into longitudinal and transverse streams, applies FiLM-conditioned 1-D spectral convolution, and uses cross-stream crosstalk to exchange information between the two streams. Third, in the \textit{readout} stage (Section~3.5), the refined directional latent sequences are decoded through spectral and bypass branches into spatial basis matrices, which are then combined by a low-rank outer product decoder to reconstruct the output field. Throughout this section, $d$ denotes the hidden dimension, $S$ is the boundary sequence length, $K$ is the low-rank readout size, and $T$ is the number of iterative message passing layers.

\subsection{Physics-Guided Edge State Initialization}
Before iterative message passing, we initialize the edge states carried by \textit{edge} nodes, which represent physical plate edges and shared interfaces (Fig.~\ref{fig:architecture}). Each \textit{edge} node is represented as a sequence of $S$ discrete sampling points along the boundary, enabling the network to handle spatially varying boundary kinematics. The latent sequence combines edge type information, positional information along the boundary, and prescribed boundary kinematic values. Edge type information specifies whether the edge is a free boundary, a boundary edge with prescribed kinematics, or an interface between adjacent plates. Positional information distinguishes sampling locations along the edge. Prescribed boundary kinematic values inject the boundary degrees of freedom into the corresponding spatial locations.

Accordingly, each \textit{edge} node $e\in\mathcal{V}_{edge}$ carries an initial sequence state $E_e^{(0)}\in\mathbb{R}^{S\times d}$ composed of three additive terms:
\begin{equation}\label{eq:edge_init}
E_e^{(0)}=B_e+P+D_e.
\end{equation}
The edge type term encodes the same edge feature vector at every sequence position:
\begin{equation}\label{eq:static_edge}
B_e=\mathbf{1}\,\phi_{edge}(x_e^{edge})^{\top},
\end{equation}
where $\mathbf{1}\in\mathbb{R}^{S}$ is the all-ones vector and $\phi_{edge}$ is a learnable embedding network. The sinusoidal positional encoding $P\in\mathbb{R}^{S\times d}$ assigns a positional representation to each sequence position:
\begin{equation}
\begin{aligned}
P[s,2k] &= \sin\left(\frac{s}{10000^{2k/d}}\right), \\
P[s,2k+1] &= \cos\left(\frac{s}{10000^{2k/d}}\right),
\end{aligned}
\end{equation}
where $k \in \{0, 1, \dots, d/2 - 1\}$ is the channel index, assuming an even hidden dimension $d$. This encoding allows the network to distinguish individual sampling points along the edge and exploit their relative spatial order during subsequent processing. The boundary kinematic term aggregates prescribed boundary values from connected \textit{boundary} nodes:
\begin{equation}\label{eq:disp_inject}
D_e[s]=\sum_{b\in\mathcal{N}_{bc}(e)}\phi_{bc}(x_b^{bc}[s]),
\end{equation}
where $\phi_{bc}$ is a pointwise projection network. For \textit{edge} nodes that do not receive any boundary kinematic input, this term is set to $D_e=\mathbf{0}\in\mathbb{R}^{S\times d}$.

\subsection{Physics Conditioning and FiLM Modulation}
For each \textit{geo} node $i$, the geometry and loading features are embedded into the hidden space:
\begin{equation}\label{eq:geo_load_embed}
\begin{aligned}
h_i^{geo}&=\phi_{geo}(x_i^{geo}), \\
h_i^{load}&=\phi_{load}(x_i^{load}),
\end{aligned}
\end{equation}
where $\phi_{geo}(\cdot)$ and $\phi_{load}(\cdot)$ are learnable embedding networks mapping to $\mathbb{R}^{d}$. These conditioning features are computed once and reused across all $T$ iterative dual-stream message passing layers.

At each message passing layer $t$, the model aggregates edge states into each \textit{geo} node. Because edge states associated with the longitudinal and transverse directions represent different directional field information, we group the relation types into a longitudinal set $\mathcal{R}_x$ and a transverse set $\mathcal{R}_y$ and aggregate them separately:
\begin{equation}\label{eq:boundary_agg}
\begin{aligned}
M_{i,x}^{(t-1)}&=\frac{1}{|\mathcal{R}_x|}\sum_{r\in\mathcal{R}_x} \sum_{e\in\mathcal{N}_r(i)} E_e^{(t-1)}, \\
M_{i,y}^{(t-1)}&=\frac{1}{|\mathcal{R}_y|}\sum_{r\in\mathcal{R}_y} \sum_{e\in\mathcal{N}_r(i)} E_e^{(t-1)}.
\end{aligned}
\end{equation}
This aggregation preserves the distinction between the two directional groups while keeping the relation specification general.

Plate geometry and applied loading influence the magnitude and spatial distribution of the stress and displacement fields. We therefore use them as conditioning factors to modulate the aggregated edge state sequences, rather than concatenating them with the edge features. This conditioning is implemented using Feature-wise Linear Modulation (FiLM) \cite{perez2018film} (Fig.~\ref{fig:architecture}):
\begin{equation}\label{eq:film_params}
[\gamma_i,\beta_i]=\phi_{cond}([h_i^{geo},h_i^{load}]),
\end{equation}
\begin{equation}\label{eq:film_modulation}
\widetilde{M}_{i,\alpha}^{(t-1)}[s]=M_{i,\alpha}^{(t-1)}[s]\odot(1+\gamma_i)+\beta_i,\qquad\alpha\in\{x,y\},
\end{equation}
where $\phi_{cond}(\cdot)$ generates the scaling vector $\gamma_i\in\mathbb{R}^{d}$ and the shifting vector $\beta_i\in\mathbb{R}^{d}$ from the concatenated conditioning features. This pointwise affine transformation adapts the aggregated edge state sequences to the geometry and loading of each plate before the dual-stream spectral convolution.

\subsection{Dual-Stream Message Passing and Crosstalk}
Stress and displacement fields on a plate component can vary differently along the longitudinal and transverse directions. We therefore keep the aggregated edge state sequences in two directional groups before the spectral update. For each plate component (\textit{geo} node), including a base plate strip, stiffener web, or stiffener flange, the longitudinal stream processes information along the longer in-plane direction. The transverse stream processes information along the perpendicular in-plane direction. Keeping the two directional sequences separate helps preserve direction dependent field variations before cross-stream information exchange. The modulated edge state sequences $\widetilde{M}_{i,x}$ and $\widetilde{M}_{i,y}$ are processed independently through two parallel 1-D spectral convolution blocks (Fig.~\ref{fig:architecture}):
\begin{equation}\label{eq:dual_stream}
\begin{aligned}
\bar{X}_i^{(t)}&=f_x\big(\widetilde{M}_{i,x}^{(t-1)}\big), \\
\bar{Y}_i^{(t)}&=f_y\big(\widetilde{M}_{i,y}^{(t-1)}\big),
\end{aligned}
\end{equation}
where $f_x(\cdot)$ and $f_y(\cdot)$ are 1-D spectral convolution blocks with the same architecture but separate parameters, allowing each stream to learn spectral representations for one plate direction.

Although the two streams are processed separately, stress and displacement fields on a plate are two-dimensional and remain coupled across the two directions. To exchange information between the two streams without fully merging the directional sequences, we introduce a cross-stream crosstalk step. Each stream is first summarized by average pooling over the sequence dimension:
\begin{equation}\label{eq:crosstalk_pool}
\begin{aligned}
p_{x,i}^{(t)}&=\frac{1}{S}\sum_{s=1}^{S}\bar{X}_i^{(t)}[s], \\
p_{y,i}^{(t)}&=\frac{1}{S}\sum_{s=1}^{S}\bar{Y}_i^{(t)}[s].
\end{aligned}
\end{equation}
The pooled summary from each stream is then linearly transformed and added as a residual to the opposite stream:
\begin{equation}\label{eq:crosstalk_inject}
\begin{aligned}
S_{x,i}^{(t)}&=\bar{X}_i^{(t)}+\mathbf{1}\big(W_{yx}p_{y,i}^{(t)}+b_{yx}\big)^{\top}, \\
S_{y,i}^{(t)}&=\bar{Y}_i^{(t)}+\mathbf{1}\big(W_{xy}p_{x,i}^{(t)}+b_{xy}\big)^{\top},
\end{aligned}
\end{equation}
where $W_{yx},W_{xy}\in\mathbb{R}^{d\times d}$ and $b_{yx},b_{xy}\in\mathbb{R}^{d}$ are learnable parameters, and $\mathbf{1}\in\mathbb{R}^{S}$ is the all-ones vector that replicates the transformed summary at every sequence position. This mechanism allows each stream to incorporate a summary from the other stream at every iteration while keeping the sequence length and ordering unchanged.

\subsection{Edge State Aggregation and Update}
After the dual-stream update, the refined directional latent sequences of each \textit{geo} node are projected back to connected \textit{edge} nodes through relation aware projections:
\begin{equation}\label{eq:edge_proj}
Q_{i,r}^{(t)}=P_r\big(S_{\kappa(r),i}^{(t)}\big),
\end{equation}
where $P_r(\cdot)$ is a relation specific projection and $\kappa(r)\in\{x,y\}$ maps relation type $r$ to the corresponding stream.

Incoming messages from \textit{geo} nodes are then aggregated at each \textit{edge} node:
\begin{equation}\label{eq:edge_agg}
A_e^{(t)}=\sum_{r\in\mathcal{R}}\sum_{i\in\mathcal{N}_r^{-1}(e)}Q_{i,r}^{(t)},
\end{equation}
where $\mathcal{N}_r^{-1}(e)$ is the set of \textit{geo} nodes that send messages to \textit{edge} node $e$ through relation $r$. The edge state is updated with a scaled residual connection followed by layer normalization:
\begin{equation}\label{eq:edge_update}
E_e^{(t)}=\mathrm{LN}\!\left(E_e^{(t-1)}+\eta A_e^{(t)}\right),
\end{equation}
where $\eta$ is the residual step scale. When multiple plate components share the same \textit{edge} node, their messages are combined in the same edge state. This allows information to pass across shared interfaces without an explicit interface consistency loss.

The complete iterative message passing procedure is summarized in Algorithm~\ref{alg:message_passing}.

\begin{algorithm}[!htbp]
\caption{DS-HGNN Iterative Dual-Stream Message Passing}
\label{alg:message_passing}
\begin{algorithmic}[1]
\Require Heterogeneous graph $\mathcal{G}$ with node features $x_e^{edge}$, $x_b^{bc}$, $x_i^{geo}$, $x_i^{load}$; iterative message passing layers $T$; step scale $\eta$
\Ensure Refined directional latent sequences $S_{x,i}^{(T)}$, $S_{y,i}^{(T)}$ for all $i \in \mathcal{V}_{geo}$
\Statex \textit{\% Initialization}
\For{each \textit{edge} node $e \in \mathcal{V}_{edge}$}
    \State Compute edge feature, position encoding, and boundary kinematic values \Comment{Eqs.~\ref{eq:static_edge}--\ref{eq:disp_inject}}
    \State $E_e^{(0)} \gets B_e + P + D_e$ \Comment{Eq.~\ref{eq:edge_init}}
\EndFor
\For{each \textit{geo} node $i \in \mathcal{V}_{geo}$}
    \State Compute FiLM parameters $[\gamma_i,\,\beta_i] \gets \phi_{cond}([h_i^{geo},\,h_i^{load}])$ \Comment{Eqs.~\ref{eq:geo_load_embed}--\ref{eq:film_params}}
\EndFor
\Statex \textit{\% Iterative message passing}
\For{$t = 1$ \textbf{to} $T$}
    \For{each \textit{geo} node $i \in \mathcal{V}_{geo}$}
        \State Aggregate edge state sequences $M_{i,x}^{(t-1)}$, $M_{i,y}^{(t-1)}$ from connected \textit{edge} nodes \Comment{Eq.~\ref{eq:boundary_agg}}
        \State Apply FiLM modulation $\widetilde{M}_{i,\alpha}^{(t-1)} \gets M_{i,\alpha}^{(t-1)} \odot (1+\gamma_i) + \beta_i$ \Comment{Eq.~\ref{eq:film_modulation}}
        \State Process each stream via 1-D spectral convolution \Comment{Eq.~\ref{eq:dual_stream}}
        \State \quad $\bar{X}_i^{(t)} \gets f_x(\widetilde{M}_{i,x}^{(t-1)})$, \quad $\bar{Y}_i^{(t)} \gets f_y(\widetilde{M}_{i,y}^{(t-1)})$
        \State Compute cross-stream summaries by mean pooling \Comment{Eq.~\ref{eq:crosstalk_pool}}
        \State \quad $p_{x,i}^{(t)} \gets \frac{1}{S}\sum_{s=1}^{S}\bar{X}_i^{(t)}[s]$, \quad $p_{y,i}^{(t)} \gets \frac{1}{S}\sum_{s=1}^{S}\bar{Y}_i^{(t)}[s]$
        \State Add crosstalk residuals to each stream \Comment{Eq.~\ref{eq:crosstalk_inject}}
        \State \quad $S_{x,i}^{(t)} \gets \bar{X}_i^{(t)} + \mathbf{1}(W_{yx}\,p_{y,i}^{(t)}+b_{yx})^{\top}$
        \State \quad $S_{y,i}^{(t)} \gets \bar{Y}_i^{(t)} + \mathbf{1}(W_{xy}\,p_{x,i}^{(t)}+b_{xy})^{\top}$
    \EndFor
    \For{each \textit{edge} node $e \in \mathcal{V}_{edge}$}
        \State Compute incoming projections $Q_{i,r}^{(t)}$ for connected \textit{geo} nodes and relations \Comment{Eq.~\ref{eq:edge_proj}}
        \State Aggregate incoming messages $A_e^{(t)}$ \Comment{Eq.~\ref{eq:edge_agg}}
        \State Update edge state $E_e^{(t)} \gets \mathrm{LN}(E_e^{(t-1)} + \eta\, A_e^{(t)})$ \Comment{Eq.~\ref{eq:edge_update}}
    \EndFor
\EndFor
\end{algorithmic}
\end{algorithm}

\subsection{Spectral--Bypass Readout and Field Reconstruction}\label{sec:spectral_bypass_readout}
After the $T$ iterative dual-stream message passing layers, each \textit{geo} node $i$ carries two refined latent sequences, $S_{x,i}^{(T)}, S_{y,i}^{(T)} \in \mathbb{R}^{S\times d}$, encoding directional edge state information. These sequences are then decoded into spatial basis matrices $V_{x,i}\in\mathbb{R}^{n_x\times K}$ and $V_{y,i}\in\mathbb{R}^{n_y\times K}$, where $K$ is the number of basis components retained for field reconstruction.

Field prediction requires representing smooth trends from overall plate bending and local high-gradient variations near plate intersections or stiffener edges. A single linear upsampling can blur local variations, while reconstruction based only on truncated Fourier modes may smooth local high-gradient variations. To account for both smooth trends and local high-gradient variations, we decode each sequence through two parallel branches, a spectral branch and a bypass branch (Fig.~\ref{fig:architecture}).

\subsubsection{Spectral Branch}
The following formulations are shared across both streams. For a generic sequence $Z\in\mathbb{R}^{S\times d}$ and target length $N\in\{n_x,n_y\}$, the spectral branch operates on truncated Fourier modes and maps them to the target resolution. Specifically, the sequence is transformed by a real FFT, multiplied by learnable complex weights over the first $M$ frequency modes, and projected to the target length by an inverse real FFT:
\begin{equation}\label{eq:spec_rfft}
\widehat{Z}=\mathrm{rFFT}(Z),
\end{equation}
\begin{equation}\label{eq:spec_weights}
\widehat{V}_{spec}[q,k]=\sum_{h=1}^{d}\widehat{Z}[q,h]\,W^{spec}[h,k,q],
\end{equation}
\begin{equation}\label{eq:spec_irfft}
V_{spec}(Z;N)=\mathrm{iRFFT}(\widehat{V}_{spec};N),
\end{equation}
where $q$ is the frequency mode index, $h$ is the hidden channel index, and $W^{spec}\in\mathbb{C}^{d\times K\times M}$ are learnable complex spectral weights over the retained modes.

\subsubsection{Bypass Branch}
In parallel, the bypass branch applies a pointwise linear projection followed by FFT based resampling to the target length, retaining local spatial features that may be lost through spectral truncation:
\begin{equation}\label{eq:bypass}
V_{byp}(Z;N)=\mathrm{Resample}_{fft}(ZW^{byp}+b^{byp};N),
\end{equation}
where $W^{byp}\in\mathbb{R}^{d\times K}$ and $b^{byp}\in\mathbb{R}^{K}$ are learnable parameters.

\subsubsection{Branch Fusion and Field Prediction}
The two branches are fused by summation, combining the spectral branch for smooth global patterns with the bypass branch for local high-gradient variations:
\begin{equation}\label{eq:fusion}
V(Z;N)=V_{spec}(Z;N)+V_{byp}(Z;N).
\end{equation}
Applying this to both streams yields the directional basis matrices:
\begin{equation}\label{eq:dir_bases}
\begin{aligned}
V_{x,i}&=V(S_{x,i}^{(T)};n_x),
\\
V_{y,i}&=V(S_{y,i}^{(T)};n_y).
\end{aligned}
\end{equation}

The plate field is then reconstructed by a low-rank outer product decoder:
\begin{equation}\label{eq:outer_product}
\hat{Y}_i[a,b,c]=\sum_{k=1}^{K}V_{x,i}[a,k]\,V_{y,i}[b,k]\,W^{core}[k,c]+b^{out}[c],
\end{equation}
where $a\in\{1,\ldots,n_x\}$, $b\in\{1,\ldots,n_y\}$, $c\in\{1,\ldots,C\}$, $W^{core}\in\mathbb{R}^{K\times C}$ is the channel mixing matrix, and $b^{out}\in\mathbb{R}^{C}$ is the output bias. This factorization represents the field using separated longitudinal and transverse basis matrices. The final channel mixing step remains compact because $W^{core}$ contains only $K\times C$ parameters. This design avoids a direct dense mapping from latent states to every grid point in the two dimensional field. The complete readout and field reconstruction procedure is summarized in Algorithm~\ref{alg:readout}.

\begin{algorithm}[!htbp]
\caption{Spectral--Bypass Readout and Field Reconstruction}
\label{alg:readout}
\begin{algorithmic}[1]
\Require Refined directional latent sequences $S_{x,i}^{(T)},\, S_{y,i}^{(T)} \in \mathbb{R}^{S \times d}$; target resolutions $n_x$, $n_y$; rank $K$
\Ensure Predicted field $\hat{Y}_i \in \mathbb{R}^{n_x \times n_y \times C}$ for each \textit{geo} node $i$
\For{each \textit{geo} node $i \in \mathcal{V}_{geo}$}
    \For{$(Z, N, V_{\alpha,i}) \in \{(S_{x,i}^{(T)},\, n_x,\, V_{x,i}),\; (S_{y,i}^{(T)},\, n_y,\, V_{y,i})\}$}
        \State Apply learnable complex weights to truncated Fourier modes \Comment{Eqs.~\ref{eq:spec_rfft}--\ref{eq:spec_irfft}}
        \State \quad $V_{spec} \gets \mathrm{iRFFT}\!\big(W^{spec} \cdot \mathrm{rFFT}(Z);\, N\big)$
        \State Project to rank $K$ basis space and resample to target resolution \Comment{Eq.~\ref{eq:bypass}}
        \State \quad $V_{byp} \gets \mathrm{Resample}_{fft}(Z W^{byp} + b^{byp};\, N)$
        \State Fuse branches $V \gets V_{spec} + V_{byp}$ \Comment{Eq.~\ref{eq:fusion}}
        \State Store fused output $V_{\alpha,i} \gets V$
    \EndFor
    \State Obtain directional bases $V_{x,i}$, $V_{y,i}$ \Comment{Eq.~\ref{eq:dir_bases}}
    \State Reconstruct field using low-rank outer product \Comment{Eq.~\ref{eq:outer_product}}
    \State \quad $\hat{Y}_i[a,b,c] \gets \sum_{k=1}^{K} V_{x,i}[a,k]\, V_{y,i}[b,k]\, W^{core}[k,c] + b^{out}[c]$
\EndFor
\end{algorithmic}
\end{algorithm}

\section{Data Preparation}
\begin{figure}[ht]
	\centering
	\includegraphics[width=0.7\linewidth]{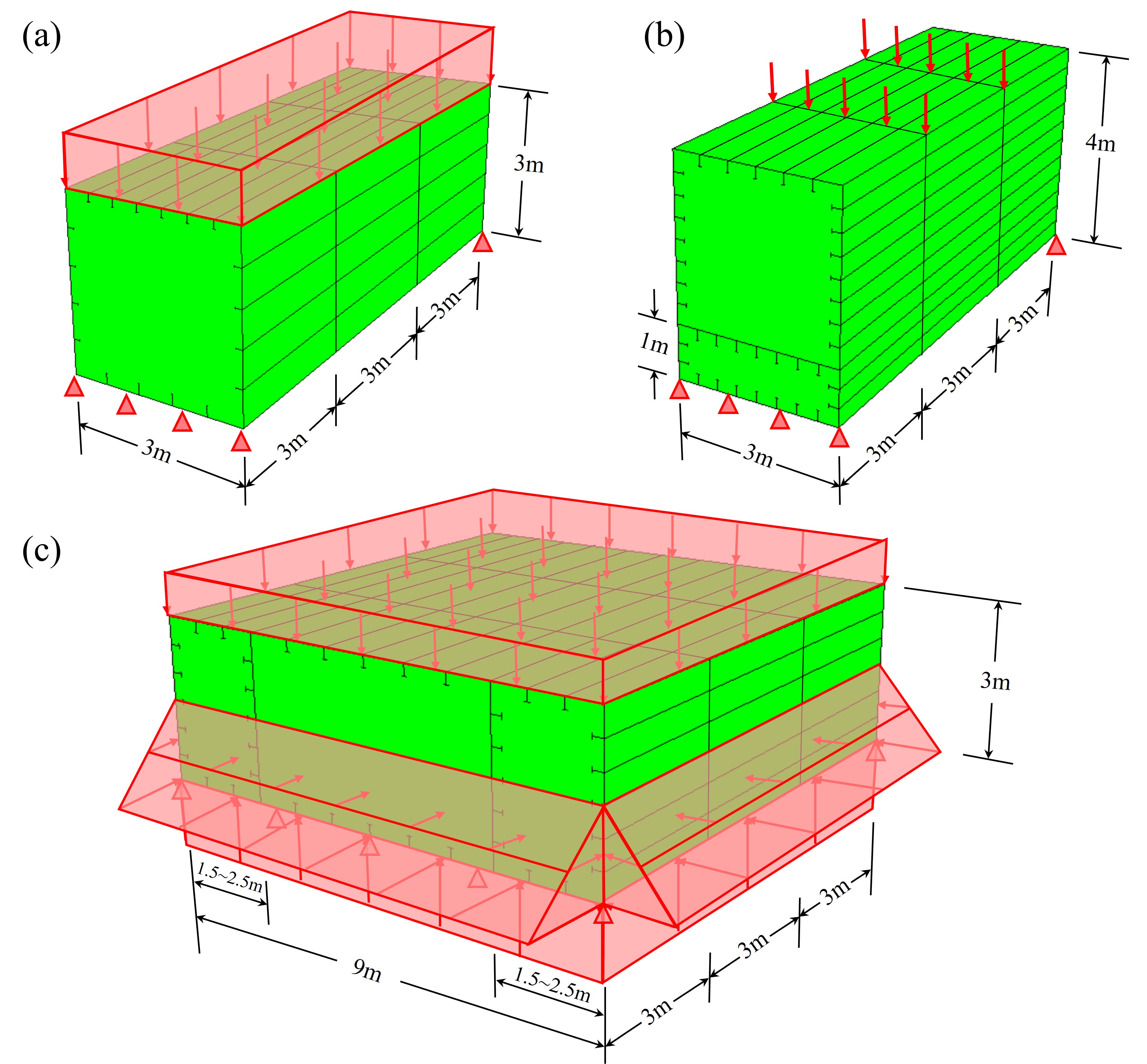}
	\caption{Geometry, loading, and boundary conditions for the three box beam cases: (a) Case~1, single-cell box beam under uniform pressure; (b) Case~2, two-cell box beam with a smaller bottom cell under four-point bending; and (c) Case~3, three-cell box beam with one center cell and two side cells under uniform and hydrostatic pressure loading.}
	\label{fig:boxbeam_configs}
\end{figure}

The dataset is constructed from three stiffened box beam cases with different structural configurations and loading conditions. These cases represent thin-walled structural systems, such as ship hulls and bridge girders, composed of different numbers of compartments. The finite element modeling and stiffened panel generation framework have been described and validated in \cite{cai2026hybrid}. The geometry, loading, and boundary conditions of all three cases are illustrated in Fig.~\ref{fig:boxbeam_configs}.

Case~1 is a single-cell box beam subjected to uniform pressure on the top panel, with pressure magnitudes ranging from $1.11\times10^5$ to $3.33\times10^5$~Pa. Case~2 is a two-cell box beam with a smaller cell located at the bottom, subjected to four-point bending applied as line loads at the intersections of transverse bulkheads and top panels, with load magnitudes ranging from 500 to 1500~kN/m. Case~3 is a three-cell box beam with one center cell and two side cells, subjected to uniform pressure on the top and bottom panels (ranging from $6.17\times10^4$ to $1.85\times10^5$~Pa) and linearly varying hydrostatic pressure on the side panels. The hydrostatic pressure acts from the bottom of the side panel to the waterline, which is determined from the structural self weight and the applied pressure. The width of the side cells varies between 1.5~m and 2.5~m. Each box beam contains four equally spaced transverse bulkheads, which are modeled as 60~mm thick isotropic plates. These transverse bulkheads provide internal supports and boundary interactions, but they are not treated as stiffened panel samples. For dataset generation, the remaining stiffened parts of the box beam structures are decomposed into panel-level samples, and each sample corresponds to one stiffened panel with its own geometry, loading, and boundary kinematics.

Because the panels are extracted from global box beam analyses, loading conditions differ across the three cases, while boundary kinematics vary from panel to panel according to panel location and global structural response. Across all cases, the extracted panels receive non-uniform boundary kinematics from the surrounding box beam structure. Not all extracted panels are directly subjected to pressure loading. In Case~1, panels on the top surface are subjected to uniform pressure, whereas panels in Case~2 have no direct pressure loading. In Case~3, panels on the top and bottom surfaces are subjected to uniform pressure, while side panels are subjected to linearly varying hydrostatic pressure. The neural network training is therefore defined at the stiffened panel level rather than at the full box beam level, where each training sample corresponds to one extracted stiffened panel. This formulation keeps the graph size moderate and avoids full box beam graph models, which could become prohibitively expensive as structural complexity increases. Stiffened panels from all three cases are randomly sampled and merged into one dataset to improve diversity. Table~\ref{tab:var_limit} summarizes the ranges of geometric variables for the stiffened panels across all three cases.

\begin{table}[!htbp]
\centering
\small
\caption{Lower and upper limits of geometric variables for stiffened panels in the dataset.}
\label{tab:var_limit}
	\begin{tabular}{@{}lccc@{}}
	\toprule
	Category (unit)                 & Case 1 \& 2 & Case 3\\
	\midrule
	Plate thickness (mm)          & 10–20        & 5–10\\ 
	Web thickness (mm)      & 5–20         & 4–8\\ 
	Web height (mm)         & 100–200      & 100–200\\ 
	Flange thickness (mm)        & 5–20         & 4–8\\ 
	Flange width (mm)             & 50–100       & 50–100\\ 
	Number of stiffeners    & 2–7          & 2–7\\ 
	\bottomrule
\end{tabular}
\end{table}

\begin{figure}[!htbp]
\centering
\includegraphics[width=0.55\linewidth]{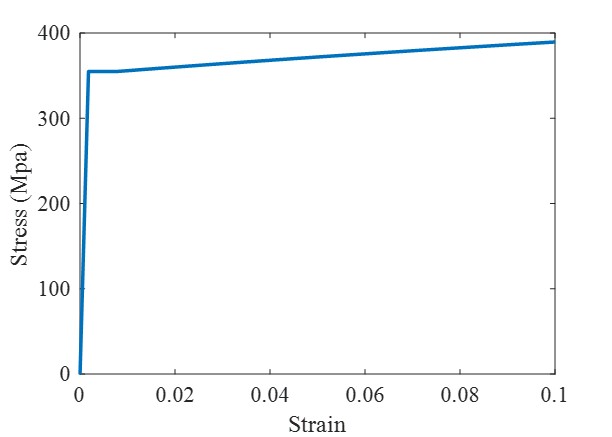}
\caption{Elastic--plastic steel stress--strain curve employed in this study.}
\label{fig:material_curve}
\end{figure}

The material model used in this study is elastic-plastic steel with Young's modulus $E=200$ GPa and Poisson's ratio $\nu=0.3$. The initial yield stress is $\sigma_0=355$ MPa, and the nonlinear plastic behavior follows:
\begin{equation}
	\sigma_f(\overline{\varepsilon}) =
	\begin{cases}
		\sigma_0 & \text{if $\overline{\varepsilon} \leq \varepsilon_L$}\\
		K(\overline{\varepsilon}_0 + \overline{\varepsilon})^n & \text{if $\overline{\varepsilon} > \varepsilon_L$}
	\end{cases}\label{Eq: Material}       
\end{equation}
where
\begin{equation}
	\overline{\varepsilon}_0 = (\sigma_0 /K)^{1/n} - \varepsilon_L ,\label{Eq: Material2}       
\end{equation}
$\overline{\varepsilon}$ is the plastic strain, and $\varepsilon_L$ is the plateau strain, taken as $0.006$ in this study. The work-hardening parameters are $K=530$ MPa and $n=0.26$ \cite{putranto2022ultimate}. The corresponding stress--strain curve is shown in Fig.~\ref{fig:material_curve}.

When converting physical panel data into heterogeneous graphs, boundary kinematics and loading functions are discretized and encoded into vectors. The boundary sequence length is set to $S=50$ sampling points per boundary edge, and the loading function is encoded using 10 points per \textit{loading} node. The output stress and displacement fields are discretized onto a $10\times50$ grid per \textit{geo} node, consistent with the elongated strip geometry of the decomposed plate units. If the FEA mesh density differs from the predefined sampling density, modified Akima (Makima) interpolation is used for resampling.

The final dataset contains $2000$ panels, with approximately equal shares of samples extracted from the three box beam cases. In this dataset, approximately $7.9\%$ of the stiffened panels contain at least one element where the von Mises stress exceeds the yield stress. All FEA data are generated using ABAQUS with S4R shell elements, using the large-displacement formulation (NLGEOM). Mesh density and convergence validation follow the procedures established in \cite{cai2026hybrid}. The dataset is partitioned into training, validation, and test subsets at a ratio of $0.8:0.1:0.1$, corresponding to 1600, 200, and 200 panels, respectively.

\section{Results and Discussion}
\subsection{Experimental Setup}
Stress and displacement fields are predicted using two separate DS-HGNN models that share the same architecture but are trained independently. Separate training allows each model to focus on the field characteristics of one target. Von Mises stress is a scalar field with steep local gradients near plate intersections and stiffener edges. Displacement is represented by three translational components and generally exhibits smoother spatial variation. Both models use the dataset split defined in Section~4. Unless otherwise stated, all results reported in Sections~5.2--5.6 are evaluated on the corresponding test set.

Before training, each target channel $c$ is standardized as
\begin{equation}
z_{c}=\frac{y_c-\mu_c}{\sigma_c},
\end{equation}
where $y_c$ and $z_c$ are the original and standardized target values in channel $c$, respectively, and $\mu_c$ and $\sigma_c$ are the mean and standard deviation for that channel. The von Mises stress channel and the three translational displacement channels are standardized separately. Three evaluation metrics are reported in this study. The standardized root mean square error ($\mathrm{RMSE}_{std}$) is the primary metric, as it enables consistent comparison across models and targets with different physical units and scales, and corresponds to the space in which the models are trained:
\begin{equation}
\mathrm{RMSE}_{std}=\sqrt{\frac{1}{N_s}\sum_{j=1}^{N_s}(\hat{z}_j-z_j)^2},
\end{equation}
where $N_s$ is the total number of scalar entries in the evaluated target fields, $\hat{z}_j$ is the predicted standardized value of the $j$th entry, and $z_j$ is the corresponding standardized FEA reference value. For displacement, the scalar entries include all three translational components.

The physical-unit root mean square error ($\mathrm{RMSE}$) is also reported, as it provides directly interpretable accuracy in engineering units and allows comparison against engineering scales such as the yield stress and expected deformation magnitudes:
\begin{equation}
\mathrm{RMSE}=\sqrt{\frac{1}{N_s}\sum_{j=1}^{N_s}(\hat{y}_j-y_j)^2},
\end{equation}
where $\hat{y}_j$ and $y_j$ are the predicted and ground-truth physical-unit values of the $j$th scalar entry, respectively.

The normalized root mean square error (NRMSE) is reported as a complementary metric that accounts for varying field magnitudes across panels with different geometries and load levels, providing a sample-wise normalized measure that is not dominated by high-amplitude samples:
\begin{equation}
\mathrm{NRMSE}=\frac{1}{N_{test}}\sum_{n=1}^{N_{test}}\frac{\mathrm{RMSE}_{n}}{\max_{k}\,|y_{n,k}|},
\end{equation}
where $N_{test}$ is the number of test samples, $\mathrm{RMSE}_{n}$ is the physical-unit root mean square error of sample $n$, $y_{n,k}$ are the physical-unit ground-truth values across all scalar entries $k$ of sample $n$, and the denominator is the maximum absolute ground-truth value in that sample. For displacement, the maximum is taken over all grid points and all three translational components.

The default DS-HGNN configuration uses hidden dimension $d=64$, boundary sequence length $S=50$, number of iterative message passing layers $T=4$, low-rank readout size $K=32$, spectral modes $M=16$, and residual step scale $\eta=0.1$. The network is trained using the Adam optimizer with an initial learning rate of $1\times10^{-3}$ and a weight decay of $1\times10^{-4}$. A step learning rate scheduler decays the learning rate by a factor of $0.5$ every $100$ epochs. The maximum number of epochs is $500$, and the batch size is $32$. The best performing checkpoint on the validation set is saved for final testing. Results over five random seeds are reported for the benchmark comparison, ablation and variant studies, and targeted evaluation of plastic material response in Sections~5.2, 5.3, and 5.6. Section~5.4 presents a single seed Optuna hyperparameter study, and Section~5.5 uses the Optuna tuned configuration for representative panel demonstrations. Model training is implemented using PyTorch and PyTorch Geometric on a computer equipped with an NVIDIA GeForce RTX 3090 GPU.

\subsection{Comparison with Benchmark Models}
\label{sec:comparison}
To evaluate the effectiveness of DS-HGNN, we compare it with six representative benchmark heterogeneous graph networks, including HGT, RGCN, HeteroConv, HAN, SeHGNN, and HINormer. These models employ several well-established HGNN mechanisms, including relation-specific message passing (RGCN~\cite{schlichtkrull2018modeling}), type-aware attention (HGT~\cite{hu2020heterogeneous}, HAN~\cite{wang2019heterogeneous}), transformer-based global aggregation (HINormer~\cite{mao2023hinormer}), pre-computed neighbor aggregation (SeHGNN~\cite{yang2023simple}), and general heterogeneous convolution (HeteroConv~\cite{fey2019fast}). All models are configured to approximately 1.4 million trainable parameters. They are evaluated in the same standardized target space using the train/validation/test split defined in Section~4.

\begin{table}[!htbp]
\centering
\scriptsize
\setlength{\tabcolsep}{2pt}
\caption{Comparison with benchmark heterogeneous-graph models (RMSE mean $\pm$ std over five seeds). RMSE is reported in standardized and physical units, and NRMSE is sample-wise normalized.}
\label{tab:benchmark_comparison}
\begin{tabular}{@{}lcccccc@{}}
\toprule
 & \multicolumn{3}{c}{Stress} & \multicolumn{3}{c}{Displacement} \\
\cmidrule(lr){2-4} \cmidrule(lr){5-7}
Model & RMSE$_{std}$ & RMSE (MPa) & NRMSE & RMSE$_{std}$ & RMSE (mm) & NRMSE \\
\midrule
DS-HGNN    & $\mathbf{0.140 \pm 0.004}$ & $\mathbf{6.89 \pm 0.19}$ & $\mathbf{0.0591 \pm 0.0013}$ & $\mathbf{0.0889 \pm 0.0011}$ & $\mathbf{1.18 \pm 0.017}$ & $\mathbf{0.0407 \pm 0.0012}$ \\
RGCN       & $0.174 \pm 0.004$          & $8.52 \pm 0.20$          & $0.0599 \pm 0.0018$ & $0.133 \pm 0.006$          & $1.72 \pm 0.11$          & $0.0544 \pm 0.0031$ \\
HINormer   & $0.193 \pm 0.013$          & $9.48 \pm 0.63$          & $0.0682 \pm 0.0040$ & $0.129 \pm 0.003$          & $1.63 \pm 0.03$          & $0.0572 \pm 0.0030$ \\
HGT        & $0.196 \pm 0.006$          & $9.62 \pm 0.30$          & $0.0713 \pm 0.0007$ & $0.158 \pm 0.008$          & $2.10 \pm 0.19$          & $0.0672 \pm 0.0038$ \\
HeteroConv & $0.211 \pm 0.012$          & $10.3 \pm 0.59$          & $0.0712 \pm 0.0018$ & $0.146 \pm 0.005$          & $1.83 \pm 0.10$          & $0.0591 \pm 0.0016$ \\
SeHGNN     & $0.222 \pm 0.002$          & $10.9 \pm 0.11$          & $0.0895 \pm 0.0037$ & $0.154 \pm 0.002$          & $2.09 \pm 0.027$          & $0.0746 \pm 0.0043$ \\
HAN        & $0.966 \pm 0.009$          & $47.3 \pm 0.44$          & $0.519 \pm 0.019$ & $1.01 \pm 0.002$          & $12.8 \pm 0.04$          & $0.459 \pm 0.018$ \\
\bottomrule
\end{tabular}
\end{table}

As shown in Table~\ref{tab:benchmark_comparison}, DS-HGNN achieves the lowest error on both targets across the three reported metrics. For stress prediction, it obtains a standardized RMSE of $0.140$, a physical-unit RMSE of $6.89$ MPa, and an NRMSE of $0.0591$. For displacement prediction, it obtains a standardized RMSE of $0.0889$, a physical-unit RMSE of $1.18$ mm, and an NRMSE of $0.0407$. Among the benchmark models, RGCN ranks second on stress, while HINormer ranks second on displacement. This ranking suggests that relation-aware propagation is useful for stress prediction, while global attention-based aggregation may be helpful for displacement prediction. HAN substantially underperforms all other benchmark models on both targets, with standardized RMSE values of $0.966$ for stress and $1.01$ for displacement. In contrast, the remaining models have standardized RMSE values below $0.23$ for stress and below $0.16$ for displacement. This may reflect the difficulty of applying HAN's attention-based aggregation scheme to this problem. The model needs to distinguish several structural relation types, including plate-edge connections, shared interfaces, loading inputs, and boundary kinematic inputs. This distinction can be challenging for a generic attention-based aggregation mechanism.

Compared with the strongest benchmark model for each target, DS-HGNN reduces the standardized stress RMSE by $19.4\%$ relative to RGCN and the standardized displacement RMSE by $31.2\%$ relative to HINormer. In physical units, the DS-HGNN stress RMSE corresponds to approximately $1.9\%$ of the material yield stress ($355$ MPa), and the displacement RMSE corresponds to approximately $0.69\%$ of the peak displacement in the dataset ($171.46$ mm). The NRMSE values show the same trend on a sample-wise normalized basis, with average errors of approximately $5.91\%$ and $4.07\%$ of the maximum field value for stress and displacement, respectively. These results indicate that the proposed physics-guided dual-stream design provides a balanced representation of relation-aware structural information and spatial field variation.

\begin{figure}[!htbp]
\centering
\begin{minipage}[t]{0.49\linewidth}
\centering
\includegraphics[width=\linewidth]{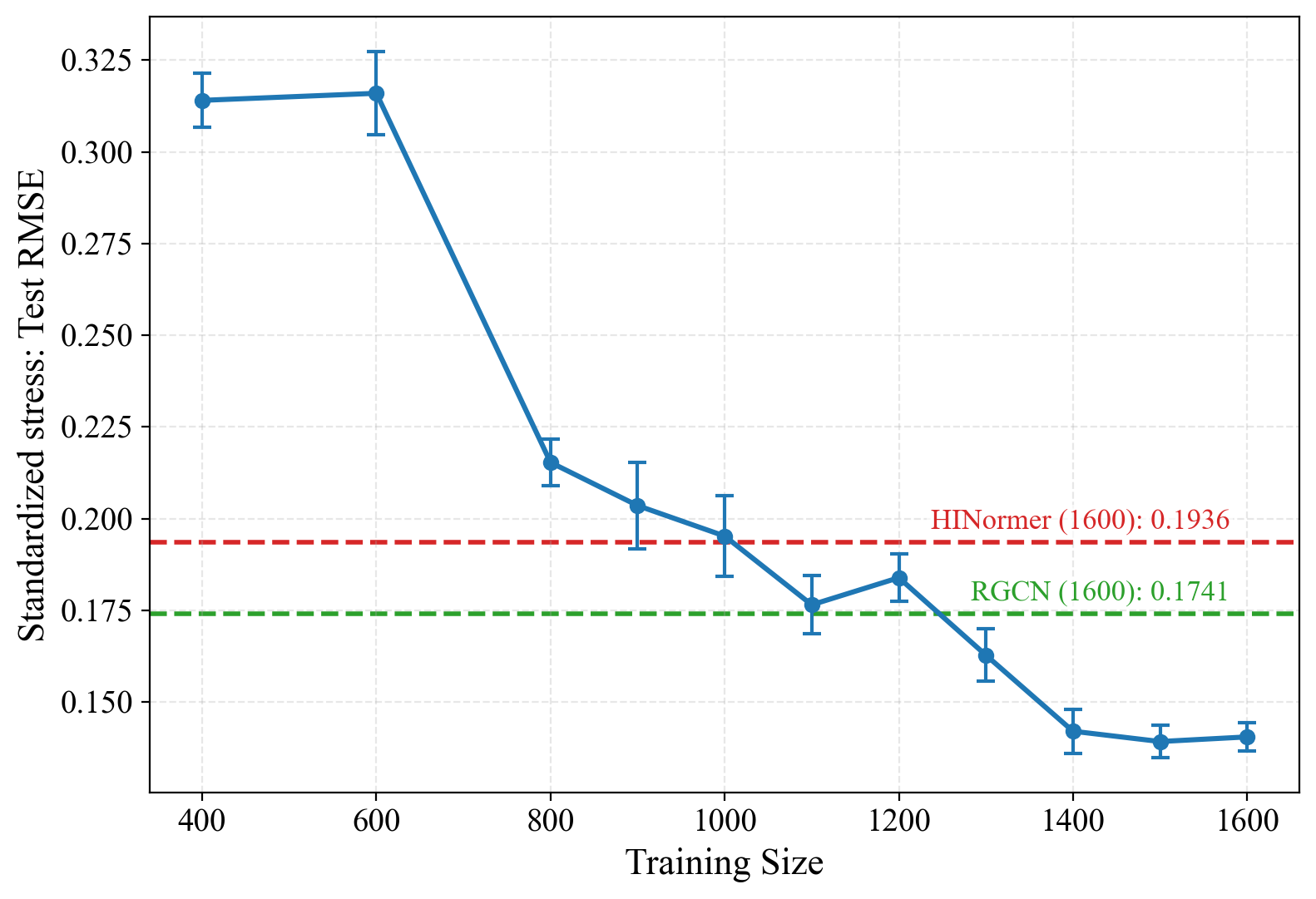}\\[-0.2em]
(a) Stress test RMSE
\end{minipage}
\hfill
\begin{minipage}[t]{0.49\linewidth}
\centering
\includegraphics[width=\linewidth]{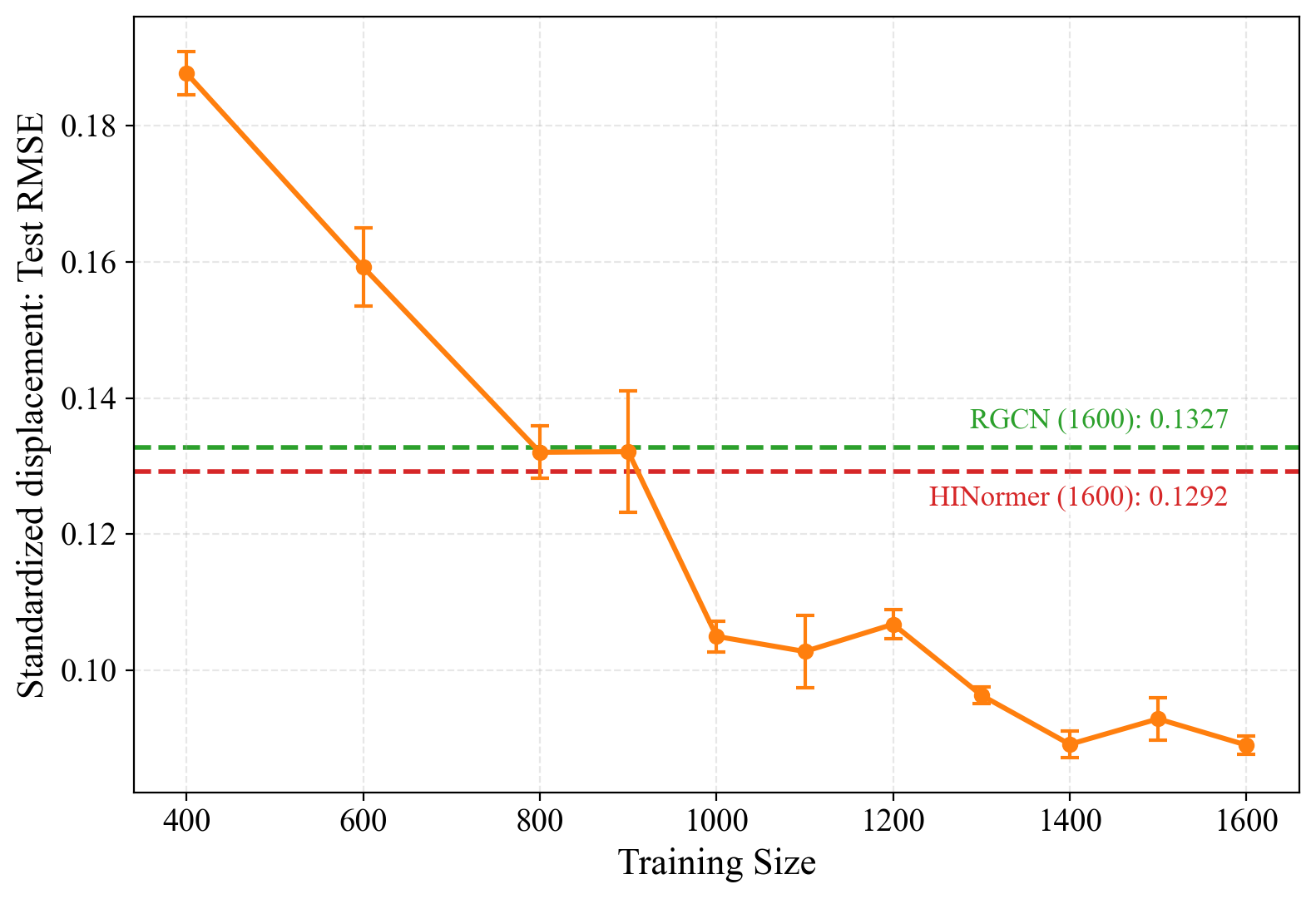}\\[-0.2em]
(b) Displacement test RMSE
\end{minipage}
\caption{Test RMSE of DS-HGNN as a function of training set size on the mixed stiffened panel dataset. Results are reported as mean$\pm$std over five seeds. In each subplot, dashed lines indicate the mean RMSE values of RGCN and HINormer trained with 1600 samples.}
\label{fig:trainsize}
\end{figure}

To assess whether this performance advantage remains when fewer training samples are available, we vary the training set size from 400 to 1600 samples while keeping validation and test sizes fixed at 200. Figure~\ref{fig:trainsize} shows the test RMSE as a function of training set size for both targets. For stress, the RMSE decreases slowly between 400 and 600 samples, decreases more rapidly between 600 and 1300 samples, and then shows smaller gains beyond approximately 1400 samples. Displacement shows a similar trend, with a clear reduction at smaller training sizes followed by smaller improvements at larger training sizes. Overall, from 400 to 1600 samples, the standardized stress RMSE decreases from $0.3141$ to $0.1405$, while the standardized displacement RMSE decreases from $0.1877$ to $0.0890$.

As shown in Fig.~\ref{fig:trainsize}, DS-HGNN matches or exceeds the accuracy of the strongest benchmark model trained with 1600 samples using fewer than 1600 training samples. For stress, RGCN achieves a standardized RMSE of $0.1741$ using 1600 training samples. DS-HGNN reaches a lower RMSE of $0.1628$ using 1300 samples, corresponding to an $18.75\%$ reduction in training samples. For displacement, HINormer achieves a standardized RMSE of $0.1292$ using 1600 samples. DS-HGNN reaches a lower RMSE of $0.1050$ using 1000 samples, corresponding to a $37.50\%$ reduction in training samples. These results indicate that DS-HGNN can achieve comparable accuracy with fewer training samples, which is important when FEA data generation is computationally expensive.

\subsection{Ablation and Variant Studies}
To quantify the contribution of each key module of DS-HGNN, we conduct ablation and variant experiments using the same data split and RMSE evaluation metric as in Section~5.2. The full model is the complete DS-HGNN with all components enabled. We evaluate six variants. Four variants remove one proposed component: the spectral branch, the bypass branch, cross-stream crosstalk, or FiLM conditioning. Two additional variants modify the readout. The Makima Readout replaces the proposed spectral and bypass basis generation with a nonparametric Makima interpolation decoder. The CNN Head Readout adds a residual two-layer CNN with $3\times3$ kernels after the proposed low-rank readout. In the Makima Readout variant, the spectral and bypass branches are replaced by a linear projection from the hidden dimension to rank $K$, followed by modified Akima interpolation to resample the projected sequences to the target grid resolutions $n_x$ and $n_y$, yielding $V_{x,i}$ and $V_{y,i}$. The low-rank outer product decoder is retained unchanged.

\begin{table}[!htbp]
\centering
\small
\caption{Ablation and variant results (RMSE mean $\pm$ std over five seeds). Relative changes are computed against the full DS-HGNN.}
\label{tab:ablation_results}
\begin{tabular}{@{}lcccc@{}}
\toprule
 & \multicolumn{2}{c}{Stress RMSE} & \multicolumn{2}{c}{Displacement RMSE} \\
\cmidrule(lr){2-3} \cmidrule(lr){4-5}
Experiment & RMSE$_{std}$ & Change & RMSE$_{std}$ & Change \\
\midrule
Full DS-HGNN           & $\mathbf{0.140 \pm 0.004}$ & $0.0\%$   & $\mathbf{0.0889 \pm 0.0011}$ & $0.0\%$   \\
No FiLM                & $0.253 \pm 0.005$          & $+80.5\%$ & $0.137 \pm 0.008$          & $+54.3\%$ \\
No Crosstalk           & $0.145 \pm 0.001$          & $+3.0\%$  & $0.102 \pm 0.004$          & $+14.7\%$ \\
No Spectral            & $0.145 \pm 0.004$          & $+3.5\%$  & $0.0927 \pm 0.0022$          & $+4.3\%$  \\
No Bypass              & $0.148 \pm 0.006$          & $+5.7\%$  & $0.103 \pm 0.002$          & $+15.8\%$ \\
Makima Readout         & $0.151 \pm 0.007$          & $+7.8\%$  & $0.098 \pm 0.004$          & $+10.2\%$ \\
CNN Head Readout       & $0.146 \pm 0.002$          & $+3.7\%$  & $0.094 \pm 0.009$          & $+5.5\%$  \\
\bottomrule
\end{tabular}
\end{table}

The results in Table~\ref{tab:ablation_results} show that the full DS-HGNN achieves the best overall performance. The largest degradation occurs when FiLM conditioning is removed, with standardized stress and displacement RMSE increasing by $80.52\%$ and $54.26\%$, respectively. This suggests that FiLM modulation of geometry and loading information is important for distinguishing panel responses under similar boundary kinematic inputs. Removing crosstalk has a relatively small impact on stress but causes a larger displacement degradation ($+14.74\%$). This trend may be explained by the fact that displacement is dominated by global plate deformation, so coupling between the two plate directions can have a larger effect on the predicted field. In contrast, the von Mises stress field contains sharper local variations near boundaries and interfaces, which may reduce the relative effect of cross-stream exchange on the aggregate RMSE.

Regarding readout design, both the No Spectral and No Bypass variants degrade performance, indicating that the spectral and bypass branches each contribute to the full model performance. Removing the bypass branch causes a larger displacement RMSE increase ($+15.83\%$) than removing the spectral branch ($+4.34\%$), indicating that the bypass branch is more important for the overall displacement accuracy in this ablation. The Makima Readout and CNN Head Readout variants also increase RMSE relative to the full model. The Makima Readout replaces the learnable spectral and bypass basis generation with a fixed interpolation scheme, which reduces the flexibility of the readout. The CNN Head Readout also gives higher RMSE, indicating that adding a residual convolutional refinement after the low-rank readout does not improve performance of the model. These results support the effectiveness of the proposed spectral--bypass low-rank decoder.

\subsection{Hyperparameter Study}
We use an Optuna-based hyperparameter optimization framework with 30 trials per target to explore the DS-HGNN hyperparameter search space. Hyperparameters are selected using standardized validation RMSE, and the corresponding standardized test RMSE is reported only after selection. All trials use a fixed seed, so the analysis focuses on trends within this controlled search rather than statistical variability across repeated runs. The search space covers low-rank readout size $K\in\{16,32,48,64\}$, Fourier modes $M\in\{8,16,24,32\}$, number of dual-stream message passing layers $T\in\{2,4,6,8\}$, learning rate (LR), and weight decay (WD).

\begin{figure}[!htbp]
\centering
\begin{minipage}[t]{0.49\linewidth}
\centering
\includegraphics[width=\linewidth]{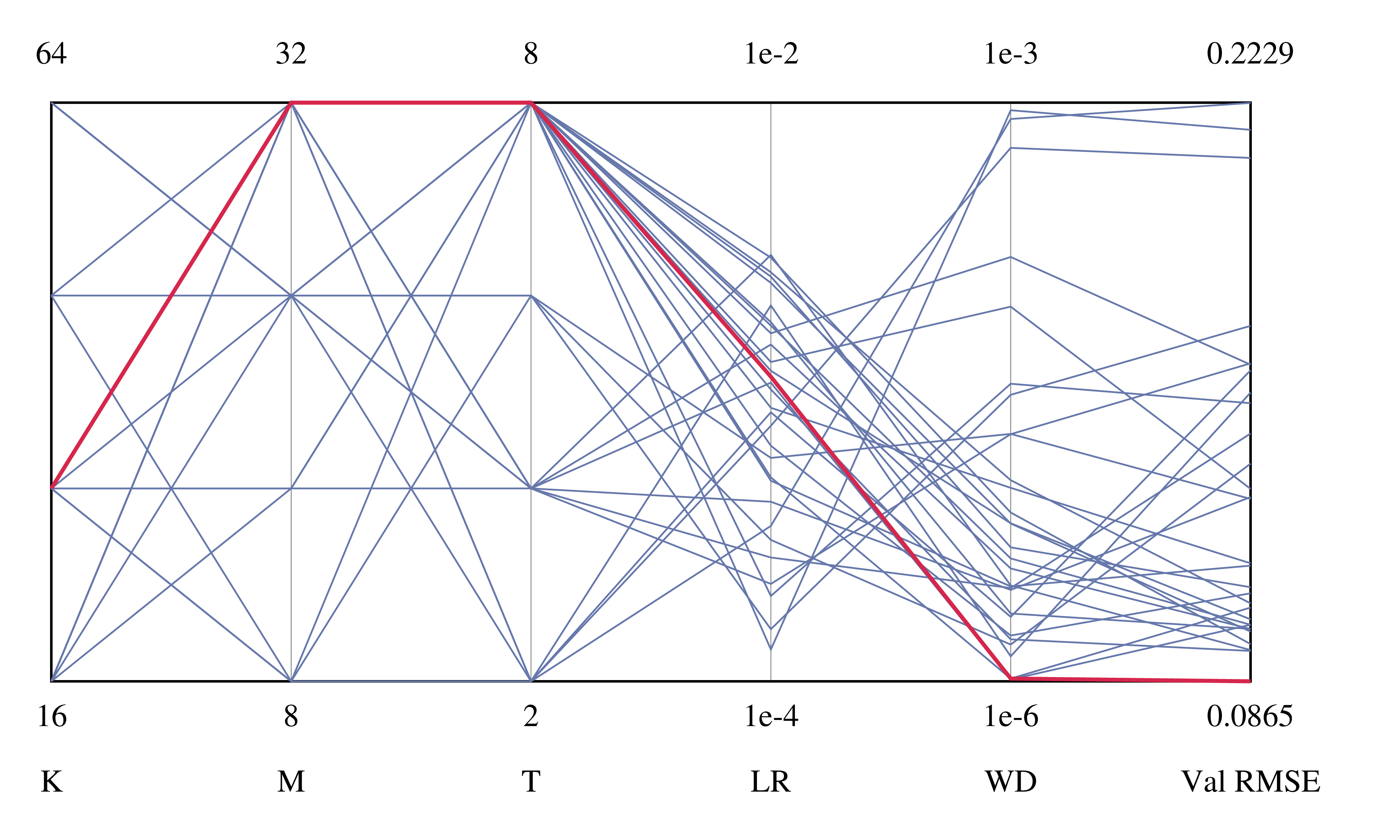}\\[-0.2em]
(a) Stress target
\end{minipage}
\hfill
\begin{minipage}[t]{0.49\linewidth}
\centering
\includegraphics[width=\linewidth]{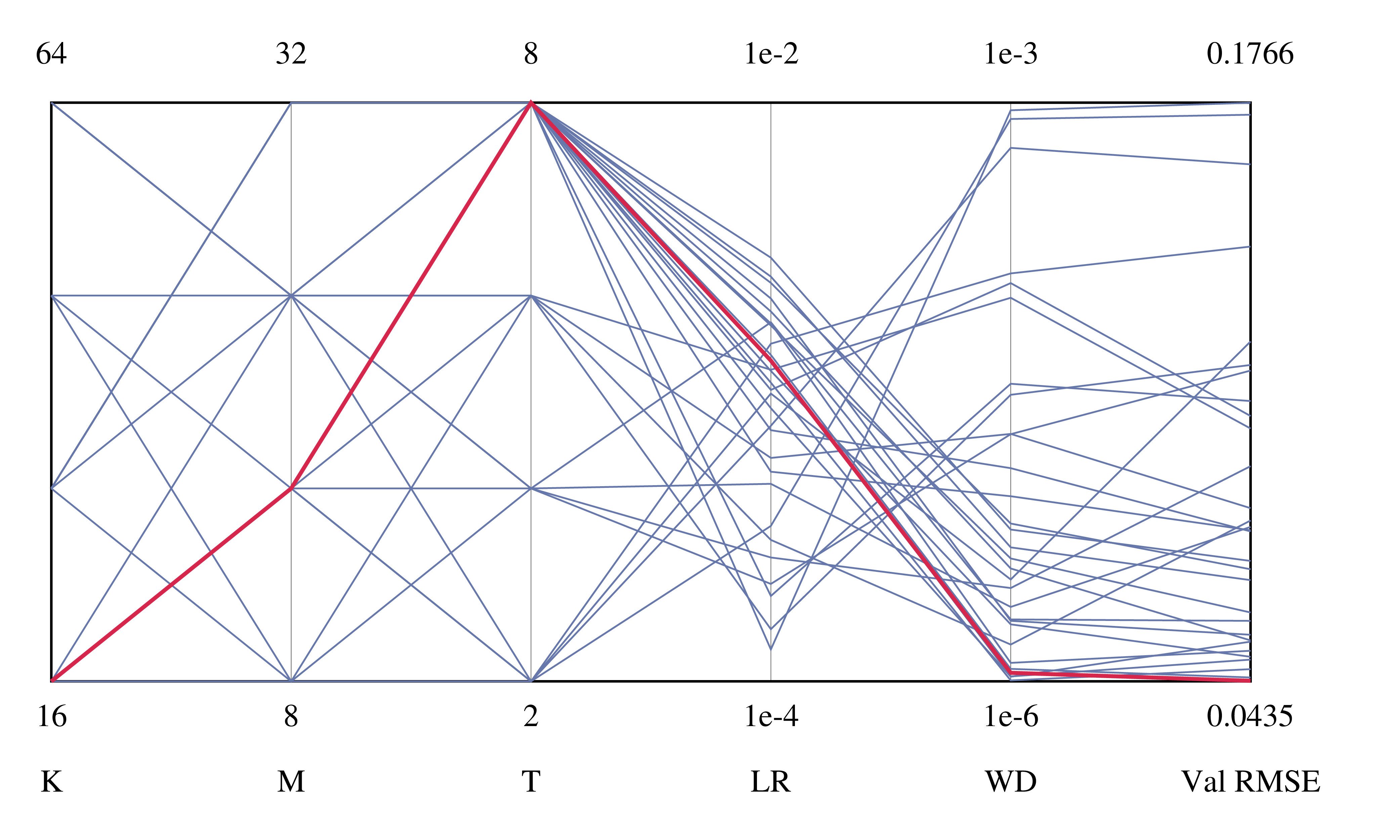}\\[-0.2em]
(b) Displacement target
\end{minipage}
\caption{Parallel coordinates plots from the Optuna hyperparameter search. Subplots show (a) stress target and (b) displacement target. Axes include low-rank readout size $K$, Fourier modes $M$, number of dual-stream message passing layers $T$, learning rate (LR), weight decay (WD), and validation RMSE.}
\label{fig:optuna_parallel}
\end{figure}

Figure~\ref{fig:optuna_parallel} shows parallel coordinates plots of all 30 trials for each target. Each axis represents one hyperparameter or the validation RMSE, and each polyline corresponds to one trial. The highlighted polyline indicates the best trial selected by validation RMSE.

The results in Fig.~\ref{fig:optuna_parallel} and Table~\ref{tab:optuna_best} show task-dependent trends. First, the selected low-rank readout size and number of Fourier modes differ between the two targets. The best stress trial uses larger values with $K=32$ and $M=32$, whereas the best displacement trial uses smaller values with $K=16$ and $M=16$. This suggests that stress prediction benefits from a larger readout basis within the tested range, whereas the smoother displacement field can be represented with fewer basis components. Second, the best trials for both targets select the largest tested number of dual-stream message passing layers ($T=8$). This suggests that a deeper neural network, with more dual-stream message passing layers, is useful within the tested search range. Larger values of $T$ were not evaluated because they would substantially increase training cost in the Optuna search. Third, the best trials for both targets use learning rates near $10^{-3}$ and very small weight decay values. Larger weight decay values are generally associated with worse validation RMSE.

\begin{table}[!htbp]
\centering
\small
\caption{Best hyperparameter configurations selected by standardized validation RMSE and their corresponding standardized test RMSE from the Optuna hyperparameter search.}
\label{tab:optuna_best}
\begin{tabular}{@{}lccccccc@{}}
\toprule
Target & Val RMSE & Test RMSE & $K$ & $M$ & $T$ & LR & WD \\
\midrule
Stress       & 0.0865 & 0.113 & 32 & 32 & 8 & $1.13\times10^{-3}$ & $1.03\times10^{-6}$ \\
Displacement & 0.0436 & 0.0570 & 16 & 16 & 8 & $1.28\times10^{-3}$ & $1.11\times10^{-6}$ \\
\bottomrule
\end{tabular}
\end{table}

Overall, the hyperparameter study suggests using target-specific values of $K$ and $M$, the largest tested number of dual-stream message passing layers ($T=8$), learning rates near $10^{-3}$, and small weight decay values within the tested range. The best trials occur near the lower bound of the weight decay search range, so this result should be interpreted as a preference for weak regularization within the tested range rather than as a fully resolved optimum. These selected configurations are used for the representative panel evaluations in Section~5.5.

\subsection{Performance on Representative Stiffened Panels}
\label{sec:panel_examples}

Using the hyperparameter configuration selected in Section~5.4, this section evaluates DS-HGNN on the three box-beam cases and presents representative field predictions for selected stiffened panels.

Table~\ref{tab:case_metrics} reports the performance of DS-HGNN for each box-beam case using standardized RMSE, physical-unit RMSE, and sample-wise NRMSE. Case~2 gives the lowest standardized and physical RMSE for both stress and displacement. In this case, the panel response is driven by boundary kinematics transferred from the global four-point bending response, without direct pressure on the panel, and the resulting fields are relatively smooth except near loaded and supported boundaries. This trend is also reflected in the displacement NRMSE, for which Case~2 gives the lowest value (0.0103). Case~1 gives the highest standardized RMSE because the uniform pressure produces large bending deformation over the top panel and increases the stress and displacement magnitudes. However, Case~1 has the lowest stress NRMSE (0.0244), indicating that its larger standardized error is mainly associated with larger field amplitudes rather than poorer proportional accuracy. Case~3 gives intermediate standardized RMSE values, but it has the highest stress and displacement NRMSE values (0.0356 and 0.0286). This suggests that the spatially varying hydrostatic pressure produces proportionally more challenging field patterns despite the moderate load amplitudes.

\begin{table}[!htbp]
\centering
\small
\caption{Performance of DS-HGNN for each box-beam case using the hyperparameter configuration selected in Section~5.4. RMSE is reported in standardized and physical units, and NRMSE is sample-wise normalized.}
\label{tab:case_metrics}
\begin{tabular}{@{}lcccccc@{}}
\toprule
 & \multicolumn{3}{c}{Stress} & \multicolumn{3}{c}{Displacement} \\
\cmidrule(lr){2-4} \cmidrule(lr){5-7}
Case & RMSE$_{std}$ & RMSE (MPa) & NRMSE & RMSE$_{std}$ & RMSE (mm) & NRMSE \\
\midrule
Case~1 & 0.162 & 7.95 & 0.0244 & 0.112 & 1.25 & 0.0177 \\
Case~2 & 0.0310 & 1.52 & 0.0312 & 0.0334 & 0.190 & 0.0103 \\
Case~3 & 0.0355 & 1.74 & 0.0356 & 0.0339 & 0.318 & 0.0286 \\
\bottomrule
\end{tabular}
\end{table}

To further examine DS-HGNN performance in predicting stress and displacement fields on stiffened panels, we select one representative panel from each case. In each case, the selected panel is the closest-to-median example in terms of overall prediction error. This choice is intended to show typical predictions rather than best-accuracy examples. The panel selected from Case~1 is subjected to uniform pressure on the top surface together with non-uniform boundary kinematics. Its maximum stress reaches the material yield limit ($355$~MPa) at the stiffener edge, so material nonlinearity is present. The panel selected from Case~2 is subjected only to non-uniform boundary kinematics induced by four-point bending of the box beam, without direct external pressure. The panel selected from Case~3 is subjected to non-uniform boundary kinematics and non-uniform hydrostatic pressure.

\begin{figure}[!htbp]
\centering
\includegraphics[width=\linewidth]{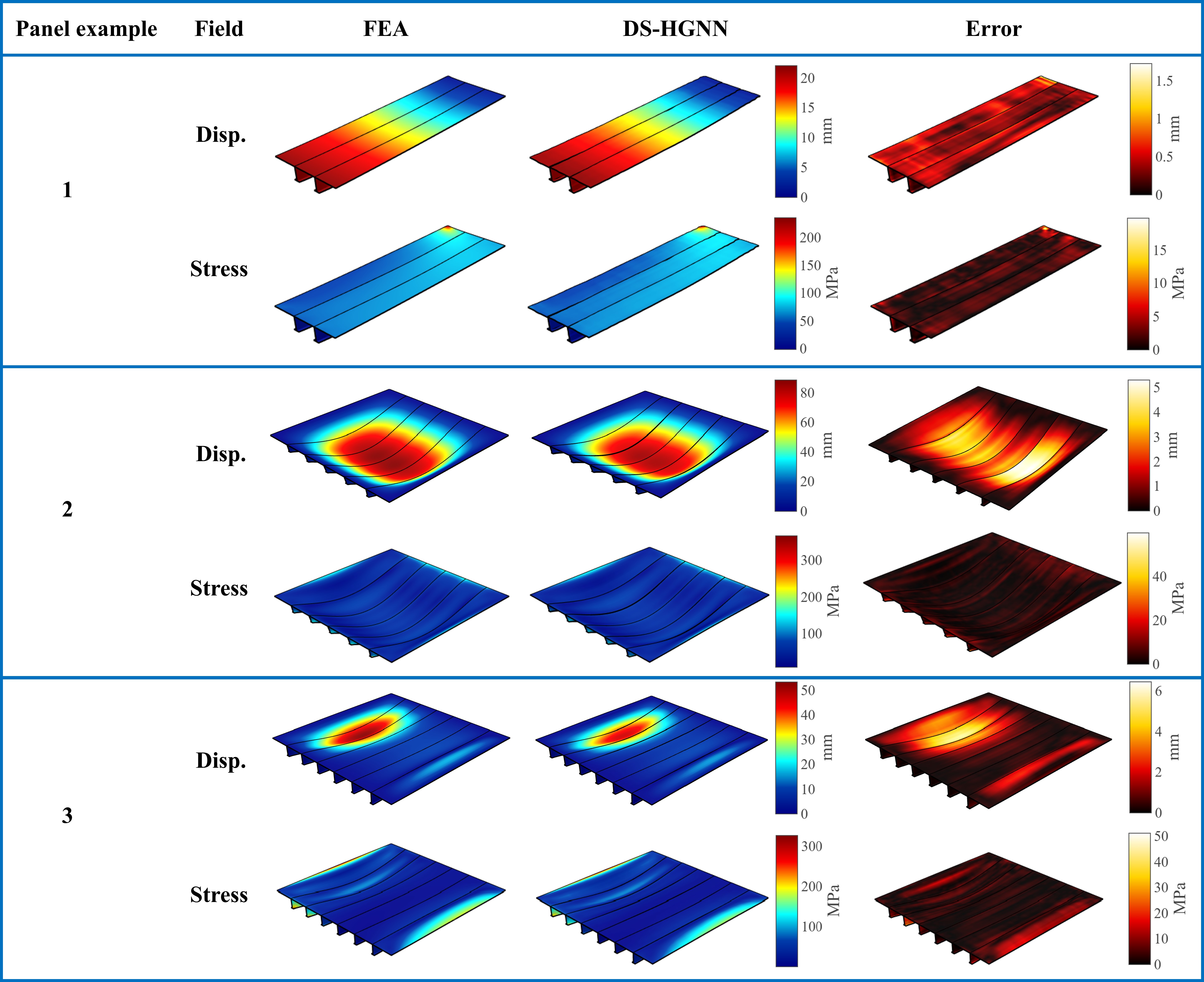}
\caption{Comparison of the total displacement and von Mises stress fields predicted by the DS-HGNN against the FEA reference for three representative panels: (a) Panel example~1 (no external pressure), (b) Panel example~2 (uniform pressure), and (c) Panel example~3 (non-uniform pressure). Each subfigure shows, from left to right, the FEA ground truth, the DS-HGNN prediction, and the absolute error map.}
\label{fig:field_comparison}
\end{figure}

\begin{figure}[!htbp]
\centering
\includegraphics[width=0.9\linewidth]{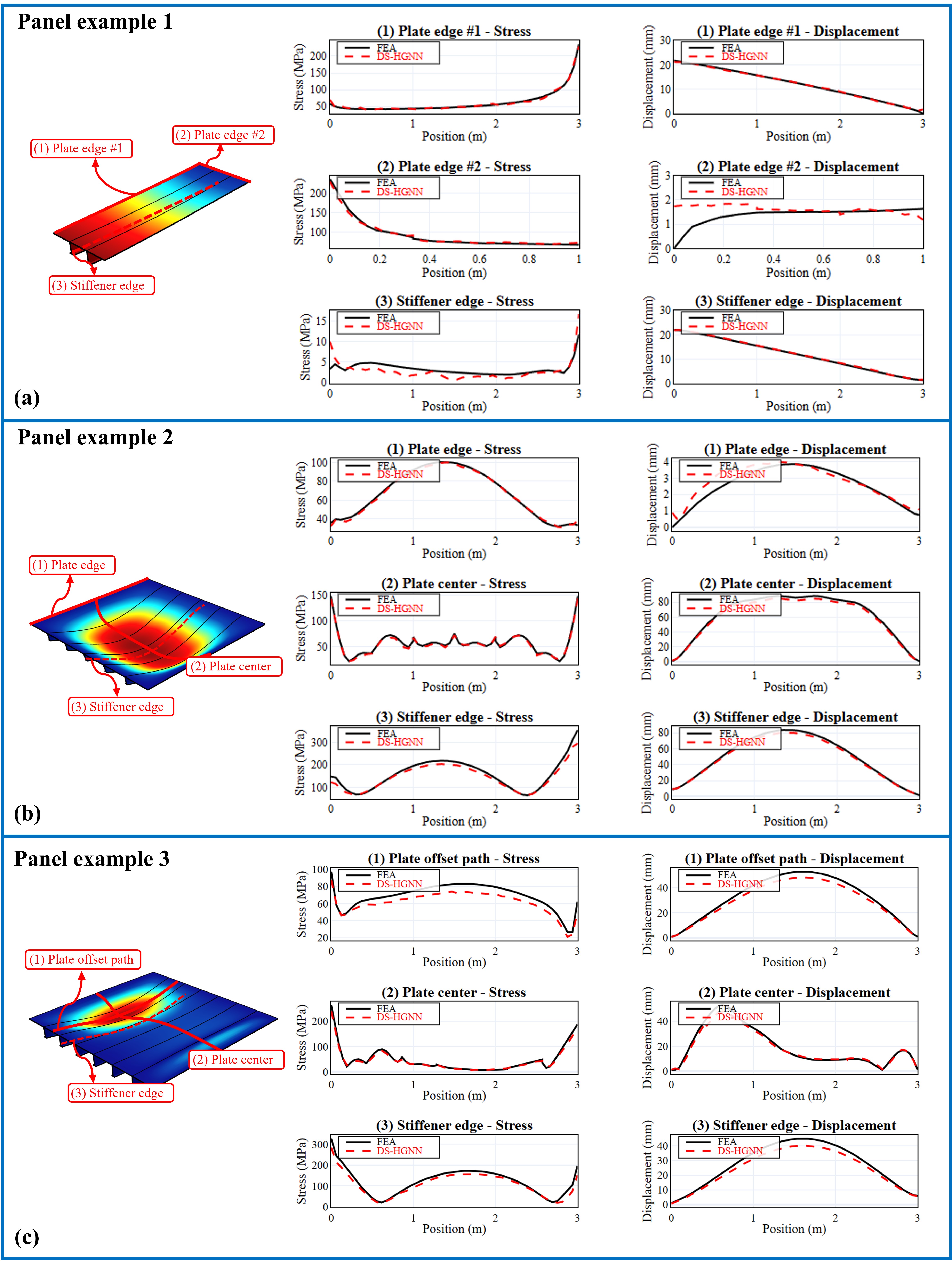}
\caption{Quantitative line comparisons of the stress and displacement profiles along three selected paths for the representative panels: (a) Panel example~1, (b) Panel example~2, and (c) Panel example~3. Black solid lines denote FEA ground truth; red dashed lines denote the DS-HGNN prediction. The inset diagram illustrates the path locations on each panel.}
\label{fig:line_comparison}
\end{figure}

In Figs.~\ref{fig:field_comparison} and~\ref{fig:line_comparison}, the selected panels are ordered by increasing loading complexity rather than by case number. Going from top to bottom in the figures, the selected panels are from Cases~2, 1, and 3, respectively. Figure~\ref{fig:field_comparison} shows the full field comparisons, and Fig.~\ref{fig:line_comparison} shows comparisons along selected paths.

In panel example~1, no direct external pressure is applied to the panel. Its response is mainly driven by boundary kinematics transferred from the global four-point bending response of the box beam. The displacement field is smooth and follows the overall bending mode. The stress field shows a local peak near the panel corner, where the support reaction causes a local stress increase. Away from this region, the stress field remains relatively smooth. DS-HGNN reproduces these main features. The maximum displacement is predicted as $21.42$~mm, compared with the FEA value of $22.04$~mm ($97.18\%$ accuracy). The local stress peak is predicted as $229.5$~MPa, compared with the FEA value of $236.1$~MPa ($97.21\%$ accuracy). The path comparisons in Fig.~\ref{fig:line_comparison}(a) show that the predicted stress and displacement profiles generally follow the FEA profiles. Most selected paths have accuracies above $95\%$. The main exception is the displacement along plate edge~\#2, for which the accuracy is $77\%$. This path has small displacement magnitudes, so a modest absolute error leads to a relatively large relative error.

Panel example~2 is subjected to uniform pressure and non-uniform boundary kinematics. The displacement field shows bending deformation over the entire panel, with the maximum displacement located near the panel center. DS-HGNN reproduces the main displacement pattern, including the peak deflection region and the low displacement regions near the panel boundaries. The stress field is more difficult to predict because material nonlinearity develops near the stiffener edge and the web--flange junction. The FEA reference shows that these regions reach the yield plateau and develop plastic response. DS-HGNN identifies the main high stress region, but it underpredicts the stress magnitude near the yielded region. At the location of the FEA stress peak, where the von Mises stress reaches $355.0$~MPa, DS-HGNN predicts $301.7$~MPa ($85.00\%$ accuracy). The path comparisons in Fig.~\ref{fig:line_comparison}(b) show the same trend. Along the plate-center path, the model follows the parabolic displacement profile produced by the uniform pressure, with $92.12\%$ displacement accuracy. Along the stiffener-edge path, DS-HGNN follows the main stress variation, but the local peak stress remains more difficult to predict.

In panel example~3, the panel is subjected to non-uniform pressure together with non-uniform boundary kinematics. The displacement field shows an asymmetric deflection pattern induced by the spatially varying pressure. DS-HGNN captures this asymmetric displacement pattern. The von Mises stress is higher along the stiffener web--flange junction. The maximum FEA stress reaches $327.1$~MPa at the stiffener edge, where DS-HGNN predicts $283.6$~MPa ($86.70\%$ accuracy). The maximum displacement is $53.49$~mm in the FEA reference and $48.41$~mm in the DS-HGNN prediction ($90.49\%$ accuracy). These errors are consistent with the more complex field variations induced by the spatially varying pressure. The path comparisons in Fig.~\ref{fig:line_comparison}(c) provide more detail on these trends. Along the plate center path, DS-HGNN follows the main stress and displacement trends, with $88.60\%$ stress accuracy and $87.10\%$ displacement accuracy. Along the stiffener edge path, the model follows the stress variation through the local high stress region, with $89.43\%$ stress accuracy, although the local peak stress remains more difficult to predict. These results show that DS-HGNN captures the main displacement and stress patterns under non-uniform loading, while local stress peaks remain the most difficult features in this example.

\subsection{Targeted Evaluation of Plastic Structural Behavior}
\label{sec:nonlinear}

While the dataset evaluated in Sections~\ref{sec:comparison}--\ref{sec:panel_examples} includes some panels with plastic material response (approximately $7.9\%$ of the dataset), the preceding evaluations mainly assess overall performance across different geometries and loading cases. This section evaluates DS-HGNN on a dataset that contains a larger fraction of panels with plastic material response. The samples are extracted from the single unit box beam (Case 1) under uniform pressure. The dataset contains 2000 panels in total, with 1600 for training, 200 for validation, and 200 for testing, matching the dataset size and split used in the preceding evaluations.

In this dataset, $25.45\%$ of panels have a maximum von Mises stress above $355$~MPa, and $25.00\%$ have a maximum von Mises stress on the $355$~MPa yield plateau. Thus, approximately half of the panels include plastic material response, either at the yield plateau or beyond. This distribution supports a targeted evaluation of DS-HGNN on plastic structural behavior.

\begin{table}[!htbp]
\centering
\scriptsize
\setlength{\tabcolsep}{2pt}
\caption{Comparison with benchmark heterogeneous-graph models on the plastic-response dataset. Values are reported as mean $\pm$ std over five seeds. RMSE is reported in standardized and physical units, and NRMSE is sample-wise normalized. Best result in bold.}
\label{tab:nonlinear_comparison}
\begin{tabular}{@{}lcccccc@{}}
\toprule
 & \multicolumn{3}{c}{Stress} & \multicolumn{3}{c}{Displacement} \\
\cmidrule(lr){2-4} \cmidrule(lr){5-7}
Model & RMSE$_{std}$ & RMSE (MPa) & NRMSE & RMSE$_{std}$ & RMSE (mm) & NRMSE \\
\midrule
DS-HGNN    & $\mathbf{0.137 \pm 0.003}$ & $\mathbf{13.5 \pm 0.25}$ & $\mathbf{0.0350 \pm 0.0005}$ & $\mathbf{0.0947 \pm 0.0029}$ & $\mathbf{3.72 \pm 0.18}$ & $\mathbf{0.0472 \pm 0.0013}$ \\
HeteroConv & $0.181 \pm 0.007$          & $18.0 \pm 0.72$          & $0.0360 \pm 0.0007$          & $0.161 \pm 0.003$          & $7.17 \pm 0.22$          & $0.0503 \pm 0.0008$          \\
HINormer   & $0.190 \pm 0.003$          & $18.8 \pm 0.26$          & $0.0393 \pm 0.0013$          & $0.159 \pm 0.002$          & $7.27 \pm 0.13$          & $0.0519 \pm 0.0006$          \\
RGCN       & $0.195 \pm 0.007$          & $19.3 \pm 0.69$          & $0.0389 \pm 0.0039$          & $0.170 \pm 0.007$          & $7.65 \pm 0.41$          & $0.0495 \pm 0.0010$          \\
HGT        & $0.212 \pm 0.006$          & $21.0 \pm 0.54$          & $0.0459 \pm 0.0009$          & $0.190 \pm 0.007$          & $8.21 \pm 0.42$          & $0.0636 \pm 0.0005$          \\
SeHGNN     & $0.259 \pm 0.009$          & $25.6 \pm 0.88$          & $0.0467 \pm 0.0013$          & $0.215 \pm 0.003$          & $9.96 \pm 0.19$          & $0.0784 \pm 0.0020$          \\
HAN        & $0.593 \pm 0.008$          & $58.7 \pm 0.78$          & $0.191 \pm 0.002$          & $0.784 \pm 0.004$          & $26.0 \pm 0.70$          & $0.412 \pm 0.025$          \\
\bottomrule
\end{tabular}
\end{table}

As shown in Table~\ref{tab:nonlinear_comparison}, DS-HGNN achieves the lowest error for both targets across all reported metrics. For stress prediction, it obtains a standardized RMSE of $0.137$, a physical-unit RMSE of $13.5$ MPa, and an NRMSE of $0.0350$. The stress RMSE in physical units corresponds to about $3.8\%$ of the yield stress ($355$~MPa). For displacement prediction, it obtains a standardized RMSE of $0.0947$, a physical-unit RMSE of $3.72$ mm, and an NRMSE of $0.0472$.

Among the benchmark models, HeteroConv gives the second-lowest stress RMSE, while HINormer gives the second-lowest displacement RMSE in standardized target space. Relative to the strongest standardized-RMSE benchmark for each target, DS-HGNN reduces the standardized stress RMSE by $24.7\%$ and the standardized displacement RMSE by $40.5\%$. These margins are larger than those observed on the mixed dataset ($19.4\%$ and $31.2\%$), suggesting that the physics-guided inductive biases in DS-HGNN provide greater benefit when a larger fraction of panels exhibit material nonlinearity. On the sample-wise normalized NRMSE metric, DS-HGNN also remains the best model, although the margins are smaller. The NRMSE reductions are $2.8\%$ for stress relative to HeteroConv and $4.6\%$ for displacement relative to RGCN. This suggests that sample-wise normalization reduces the apparent differences among the strongest models, while DS-HGNN still gives the lowest NRMSE.

\begin{figure}[!htbp]
\centering
\includegraphics[width=0.75\linewidth]{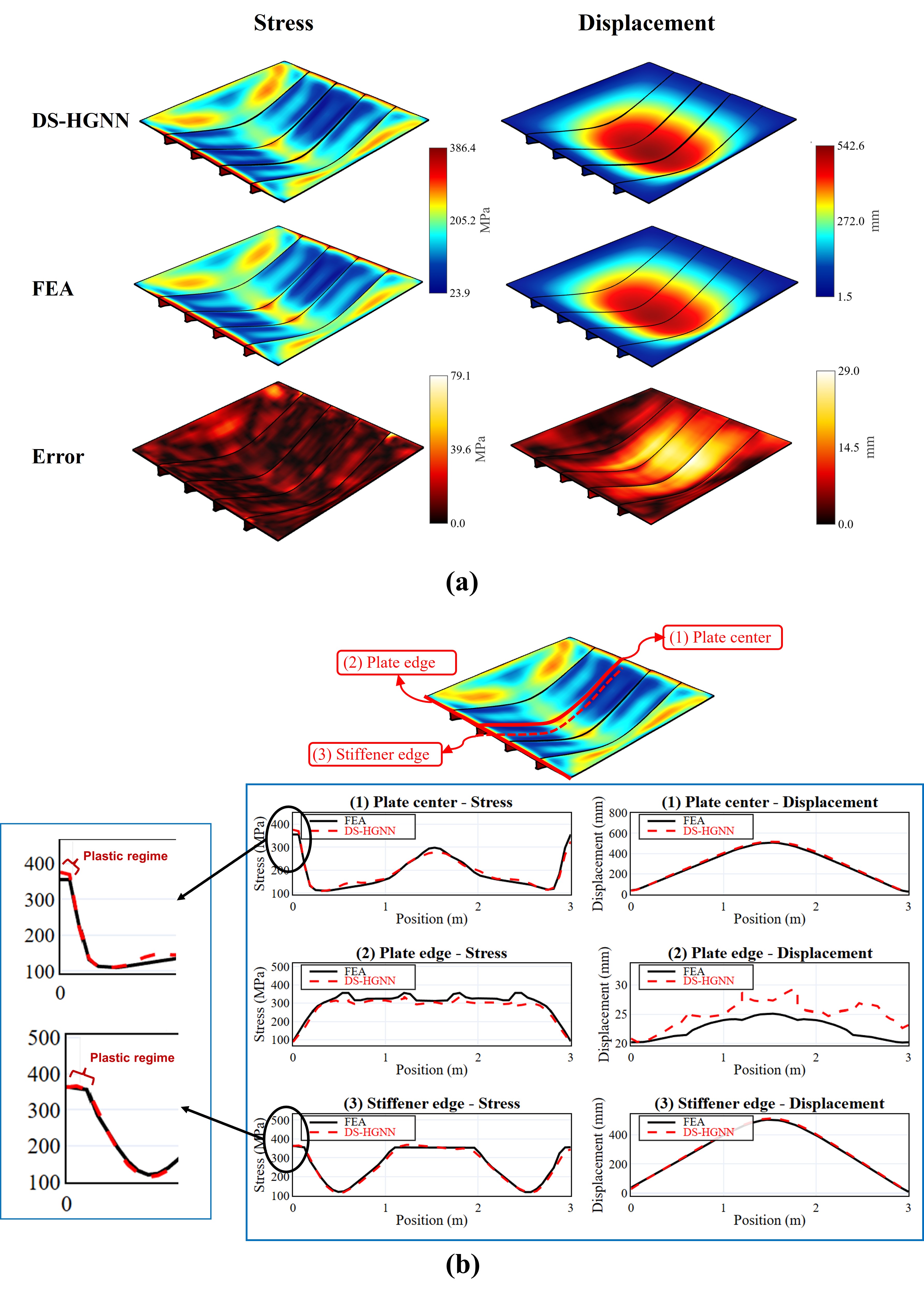}
\caption{Selected panel with nonlinear material response from the materially nonlinear dataset. (a)~Stress and displacement field comparisons between DS-HGNN predictions and FEA ground truth, together with their absolute error maps. (b)~Quantitative line comparisons of stress and displacement profiles along three selected paths: plate center, plate edge, and stiffener edge. Zoom-in insets on the plate-center and stiffener-edge paths highlight the stress plateau near the panel-bulkhead intersection, where the material has entered the plastic regime.}
\label{fig:plastic_example}
\end{figure}

Fig.~\ref{fig:plastic_example} presents the qualitative assessment for a selected panel from the test set of this targeted dataset. Unlike Section~\ref{sec:panel_examples}, where the selected panels are closest to median examples within their cases, this panel is selected to examine a clear plastic material response. Its maximum FEA von Mises stress is $364.8$~MPa, which exceeds the yield stress of $355$~MPa. Its overall prediction error is above the test set average, indicating that the example is not an unusually low error case.

The panel is subjected to uniform pressure on the top surface, with non-uniform boundary kinematics transferred from the surrounding box-beam structure. The stress field includes both elastic and plastic response regions. The von Mises stress remains in the elastic range in the central region of the plate, while plastic response develops along parts of the plate edge, the stiffener web edge, and the stiffener flange edge. The highest stress occurs at the stiffener flange near the panel-bulkhead junction, where the FEA stress rises above the yield plateau and reaches $364.8$~MPa. At this location, DS-HGNN predicts $365.9$~MPa ($99.70\%$ accuracy). The displacement field peaks at the plate center and decreases toward the panel edges, consistent with the global bending deformation under uniform pressure. The maximum FEA displacement is $528.1$~mm at the plate center, while DS-HGNN predicts $542.6$~mm ($97.26\%$ accuracy). These peak value comparisons show that the model captures the main stress and displacement magnitudes in this example.

The path comparisons in Fig.~\ref{fig:plastic_example}(b) provide more detail on the plastic response. The zoom-in insets highlight the yielded portions of the stress profiles near the panel-bulkhead intersection on the plate-center and stiffener-edge paths. In these regions, the DS-HGNN stress curves closely follow the FEA stress plateau. The middle portions of the plate-edge and stiffener-edge paths also show material plastic behavior. This demonstrates that plastic response is not limited to the panel-bulkhead intersection.

The model predicts these plastic regions with different levels of accuracy. Along the plate-edge path, DS-HGNN underpredicts parts of the interior stress plateau. This path has the largest stress error among the three selected paths, although it still maintains $92.22\%$ stress accuracy. Along the stiffener-edge path, the model follows the yielded region more closely and achieves $96.21\%$ stress accuracy, with some smoothing near the yield plateau. The corresponding displacement accuracies are $88.44\%$ and $96.55\%$ for the plate-edge and stiffener-edge paths, respectively. Along the plate-center path, the stress and displacement accuracies are $95.18\%$ and $95.62\%$, respectively. These results indicate that DS-HGNN can reproduce both smooth elastic field variations and yield plateau regions in this example. The local plateau details remain more difficult along the plate edge, where the material response shows clear nonlinearity.

\section{Conclusion and Future Work}

This paper presents the novel Dual-Stream Heterogeneous Graph Neural Network (DS-HGNN) for predicting stress and displacement fields in thin-walled structural systems, such as bridge girders, ship and offshore structures, and aerospace structures. The method is developed for structures that can be decomposed into interacting plate-like components, and is demonstrated here on stiffened panels with different geometries, boundary kinematics, and external loadings.

In this study, the proposed DS-HGNN architecture operates on stiffened panels represented as heterogeneous graphs with \textit{geo}, \textit{edge}, \textit{loading}, and \textit{boundary} nodes. The model initializes edge states from structural edge information, spatial position, and boundary kinematics, and then updates these states through iterative dual-stream message passing between plate domains and shared interfaces. This design processes longitudinal and transverse information separately while allowing information exchange between the two directions. Geometry and loading effects are included through Feature-wise Linear Modulation (FiLM) conditioning. The final stress and displacement fields are reconstructed on a regular spatial grid using the proposed spectral--bypass readout and low-rank outer product decoding.

The model was evaluated on a mixed dataset of 2000 panels from three stiffened box-beam cases. Stiffened panels from these cases cover different loading scenarios and boundary kinematics. A second dataset with a higher proportion of panels showing plastic material response was used to assess the capacity of DS-HGNN to capture yield plateau and post-yield stress features. The key findings are summarized as follows:

\begin{itemize}
    \item DS-HGNN achieves the lowest stress and displacement RMSE among the six tested benchmark heterogeneous graph neural networks.
    \item DS-HGNN matches or exceeds the accuracy of the strongest benchmark heterogeneous graph neural network using $19\%$--$38\%$ fewer training samples, indicating improved data efficiency for FEA-based surrogate modeling.
    \item The component and variant studies show that edge state initialization, dual-stream processing, cross-stream crosstalk, FiLM conditioning, and the spectral--bypass readout all contribute to the full model performance. FiLM conditioning of geometry and loading information and the spectral--bypass readout provide the largest improvements.
    \item Representative panel evaluations show that DS-HGNN captures the main stress and displacement field patterns under different loading scenarios, including cases with non-uniform boundary kinematics, uniform pressure, and non-uniform pressure.
    \item On the dataset with a higher proportion of panels showing plastic material response, DS-HGNN achieves the lowest stress and displacement RMSE among the benchmark models and reproduces both yield plateau regions and high stress regions after yielding.
\end{itemize}

In summary, DS-HGNN provides an accurate and data-efficient surrogate for full-field prediction in thin-walled structures made of stiffened panels. Future work will extend the architecture to effectively handle more severe nonlinear regimes, including large plastic deformation and buckling, as well as more complex loading environments. In addition, the architecture could be generalized to more complex structures or even solid components, where more than two directional streams may be needed to represent field variations. Semi-supervised or physics-constrained learning strategies will also be explored to reduce the reliance on labeled FEA data.

\section*{CRediT authorship contribution statement}
\textbf{Yuecheng Cai}: Conceptualization, Data curation, Formal analysis, Investigation, Methodology, Software, Validation, Visualization, Writing - original draft, Writing - review \& editing.

\noindent\textbf{Jasmin Jelovica}: Funding acquisition, Investigation, Project administration, Resources, Supervision, Writing - review \& editing.

\section*{Acknowledgments}
This research was financially supported by Seaspan Shpyards and Natural Sciences and Engineering Research Council of Canada (NSERC) through Discovery Grant RGPIN-2025-04421 and Alliance Advantage Grant ALLRP 607677-25.

\bibliographystyle{elsarticle-num}
\bibliography{Bib}

\end{document}